\documentclass[11pt,a4paper]{article}

\usepackage[a4paper,margin=2.5cm]{geometry}
\usepackage{setspace}
\usepackage[T1]{fontenc}
\usepackage[utf8]{inputenc}
\usepackage{textcomp}

\usepackage{newtxtext}
\usepackage{newtxmath}

\usepackage{graphicx}
\usepackage{booktabs}
\usepackage{array}
\usepackage[table]{xcolor}
\usepackage{enumitem}
\usepackage{threeparttable}
\usepackage{longtable}
\usepackage{caption}
\usepackage{subcaption}
\usepackage{float}
\usepackage{adjustbox}
\usepackage{makecell}
\usepackage{fancyvrb}
\usepackage{pdflscape}

\usepackage{amsmath,amssymb}

\usepackage[numbers,sort&compress]{natbib}
\usepackage[hidelinks]{hyperref}

\title{Vision Foundation Models in Radiology: A Scoping Review of Data, Methodology, Evaluation and Clinical Translation}

\author{Alejandro Vergara-Richart$^{1,2,*}$, Xavier Rafael-Palou$^{1,*}$, Almudena Fuster-Matanzo$^{1}$, \\Ignacio Iborra Roncales$^{1}$, Ángel Alberich-Bayarri$^{1}$, Ana Jiménez-Pastor$^{1}$}

\date{
$^{1}$Quantitative Imaging Biomarkers in Medicine (Quibim S.L.), Valencia, Spain\\
$^{2}$Universitat Politècnica de València, Valencia, Spain\\[0.5em]
$^{*}$These authors contributed equally to this work.\\[0.5em]
Correspondence: Alejandro Vergara-Richart, \href{mailto:alejandrovergara@quibim.com}{alejandrovergara@quibim.com}
}

\begin{document}

\maketitle

\textbf{Keywords:} foundation models; radiology; review; medical imaging 

\begin{abstract}
Vision foundation models (VFMs) are increasingly being developed for radiological imaging, yet their definition, development and evaluation remain heterogeneous. We conducted a PRISMA-ScR scoping review of peer-reviewed studies published between January 2017 and March 2026 describing foundation models trained exclusively on radiological imaging data. Sixty-seven studies were included and mapped across three pillars: data scale and heterogeneity, architectural and pretraining scalability, and downstream transferability and generalization. Datasets primarily covered brain MRI, thoracoabdominal
CT, and chest X-ray, ranging from fewer than 100,000 samples to multi-million-image cohorts. Transformer-based architectures and self-supervised pretraining predominated, particularly masked image modeling, contrastive learning and multi-stage approaches. Evaluation focused mainly on segmentation and classification, whereas cross-center, cross-scanner, anatomical and modality-shift validation was inconsistently reported. Alignment with FUTURE-AI principles was uneven. Overall, radiology-specific VFMs show promising transferability, but clinical translation remains constrained by limited data representativeness, heterogeneous benchmarks, incomplete reporting and insufficient deployment-oriented evaluation.
\end{abstract}

\section{Introduction}
\label{sec:intro}

Originating in natural language processing and popularized through large language models \citep{brown2020language}, 
foundation models (FMs) represent a shift toward learning general-purpose representations from large-scale data. Unlike task-specific models optimized for a single objective, FMs are trained on diverse and extensive datasets with the objective of capturing transferable features that can be adapted to multiple downstream applications. 

This paradigm shift has been extended to computer vision through vision foundation models (VFMs), which are high-capacity neural networks trained on large-scale, heterogeneous, and often unlabeled imaging data to learn reusable representations that can be efficiently adapted to diverse specific tasks \citep{gui2024survey}. 
These properties are particularly relevant in radiology imaging analysis, where data scarcity, high annotation costs, and substantial variability across institutions, scanners and acquisition protocols hinder predictive model generalization and clinical translation  \citep{topol2019high}. Furthermore, such variability induces distributional shifts that significantly degrade performance under external validation and real-world deployment settings.

Self-supervised learning (SSL) has become the dominant strategy for pretraining models in medical imaging due to its ability to exploit the intrinsic structure of unlabeled data \citep{de2025foundation}. In medical imaging, SSL approaches are typically grouped into two broad families: (i) reconstruction-based methods, such as masked image modeling and related generative objectives \citep{li2025self}, and (ii) contrastive learning frameworks that promote invariance across augmented views, modalities, or patient-level representations \citep{wang2023review}. Complementary, weakly supervised learning strategies, which exploit coarse or inexact labels, have also contributed to scaling representation learning in radiology \citep{misera2024weakly}.

Built upon these pretraining strategies, VFMs are typically implemented using large-capacity architectures (often based on transformer designs \citep{dosovitskiy2020image}) and trained on increasingly large and diverse datasets. Once pretrained, these models are adapted to downstream clinical tasks through a range of transfer learning strategies, including zero-shot (ZS) and few-shot (FS) inference, as well as full or partial fine-tuning \citep{fine_tuning}. More recently, parameter-efficient fine-tuning (PEFT) methods, such as low-rank adaptation \citep{lora} and adapter-based approaches \citep{peft}, have gained increasing attention. These techniques enable efficient task adaptation while reducing computational cost and mitigating overfitting, which is particularly relevant in data-scarce clinical environments.

Despite rapid progress, the concept of “vision foundation model” remains inconsistently defined across the radiology literature. Models referred to as VFMs vary substantially in dataset scale and diversity, pretraining strategy, architectural design, and downstream evaluation protocols \citep{CHEN2026100123}. This terminology ambiguity complicates systematic identification of relevant models, limits meaningful cross-study comparisons, and obscures a rigorous understanding of their strengths and weaknesses.

Several reviews have recently examined FMs in medical imaging from complementary perspectives, yet important aspects remain incompletely characterized. A recent narrative review described training paradigms, adaptation strategies, and evaluation frameworks for radiology FMs; however, it provided limited characterization of dataset composition, architectural trends, and the distribution of downstream tasks across the literature \citep{paschali2025foundation}. Another review offered a broader perspective spanning modalities beyond radiology such as pathology and ophthalmology, but at the cost of modality- and anatomy-specific characterization within radiology itself \citep{zhang2024challenges}. A survey of self-supervised pretraining examined diagnostic tasks across X-ray, CT, MRI, and ultrasound; however, its focus on a single methodological paradigm placed less emphasis on the broader landscape of FM architectures, downstream task coverage, and challenges related to clinical translation \citep{vanberlo2024survey}. A systematic review and meta-analysis of vision-language FMs quantified performance across classification, segmentation, report generation, and visual question answering, but its coverage of vision-only radiology FMs remained limited \citep{sun2025vlm}. Another recent review examined the fundamentals, applications, opportunities, challenges, risks, and prospects of FMs for radiology with particular attention to clinical adoption and regulatory considerations, although its narrative format did not provide a systematic evidence mapping of architectural choices, dataset characteristics, and external validation gaps \citep{akinci2026fundamentals}.

Taken together, these contributions highlight the growing interest in radiology foundation models, while underscoring the absence of a comprehensive and structured overview focused specifically on radiology VFMs. To address this need, we conducted a scoping review that organizes the literature around three core pillars: (i) access to large-scale and heterogeneous imaging datasets, (ii) scalable pretraining strategies and model architectures, and (iii) adaptation approaches for downstream clinical tasks. By synthesizing the literature around these interconnected pillars, this review aims to support more consistent characterization and comparison of VFMs in radiology, and to identify methodological and clinical gaps that warrant further investigation. 

Complementing this framework, we examine the extent to which included models address key principles of the FUTURE-AI guidelines \citep{futureaiguidelines}, namely robustness, fairness, universality, explainability, traceability, and usability. This perspective allows the review to situate methodological advances within broader considerations of safe, and clinically effective deployment, and to identify where current VFM development diverges from established standards for trustworthy medical AI.
\section{Methods}


A review protocol for this study was not registered.

\subsection{Eligibility criteria}
\label{subsec:eligibility_criteria}
The eligibility criteria were defined following PRISMA-ScR \citep{prismascr} recommendations to ensure transparency and reproducibility. Studies were screened according to the following inclusion criteria.

\begin{enumerate}[
    label=\textbf{\roman*.},
    leftmargin=*,
    align=left,
    nosep
]
\item Peer-reviewed journal articles in English published between January 2017 and March 2026. The starting date was selected to capture the emergence of modern large-scale representation learning paradigms and FM-like approaches in computer vision \citep{vaswani2017attention, byol}.
\item Studies involving models trained on radiological imaging data, including radiography (X-ray), computed tomography (CT), magnetic resonance imaging (MRI), nuclear medicine (PET/CT), and ultrasound (US).
\item Studies employing pretraining on imaging datasets intended to learn generalizable representations transferable across downstream tasks. This includes self-supervised learning, weakly supervised learning or supervised pretraining on large datasets.
\item Studies proposing a new VFM or reporting the adaptation and fine-tuning of an existing VFM for radiology applications.
\item Studies evaluating the VFM on at least one radiology-related downstream task, such as classification, segmentation, detection, report generation, or survival prediction.
\item Studies explicitly described by the authors as VFM were included despite partially meeting the above criteria, provided that this classification was supported through a journal peer-review process.
\end{enumerate}

\noindent Conversely, studies were excluded following the following criteria:

\begin{enumerate}[
    label=\textbf{\roman*.},
    leftmargin=*,
    nosep
]
\item Studies based exclusively on shallow learning methods or conventional machine learning approaches without deep neural networks.
\item Models developed exclusively for a single downstream task without evidence of representation, transfer, adaptation or reuse across tasks, datasets, or clinical settings.
\item Studies using fully frozen pretrained models without proposing methodological adaptations, architectural modifications, or representation learning contributions.
\item Studies focused solely on clinical application, benchmarking, or comparative evaluation without methodological development of a FM.
\item Studies proposing training pipelines or frameworks without developing a radiology VFM.
\item Studies that did not include a pretraining stage intended to learn transferable representations.
\item Studies exclusively involving non-radiological imaging domains, including histopathology, ophthalmology, microscopy, endoscopy, or dermatology imaging.
\item Editorials, commentaries, letters, conference abstracts, tutorials, surveys, and non-peer-reviewed publications.
\item Studies whose primary objective was not the development, adaptation, or evaluation of radiology-oriented VFMs.
\end{enumerate}

\vspace{0.5em}

\subsection{Information Sources and Search Strategy}
A literature search for eligible publications was conducted in March 2026 across PubMed, Scopus, IEEE, and EMBASE. The key search terms were based on a combination of two major terms: “foundation models” and “medical imaging” (see Supplementary Material Section 7.1). Search terms were formulated to include radiology-focused VFMs, while exclusion terms were applied to remove studies with multimodal, non-image-based, or non-radiology FMs.

\vspace{0.5em}

\subsubsection{Search strategy}
Full electronic search strategy for one database is provided in Supplementary Material Section 7.2.

\vspace{0.5em}

\subsubsection{Selection of Sources}
Literature search and study selection were independently conducted by two reviewers following the predefined eligibility criteria. Covidence review management software (Veritas Health Innovation, Melbourne, Australia) was used to support record management, study screening, and data extraction. After removal of duplicate records, all retrieved studies identified through the search strategy underwent a two-stage screening process consisting of (i) title and abstract screening and (ii) full-text review. Each study was independently assessed by both reviewers at both screening stages. Disagreements were resolved through consensus discussions during dedicated review meetings. 

\vspace{0.5em}

\subsection{Data charting}
Data extracted included the following: (1) general information: title, authors, affiliations, journal, publication date, and DOI; (2) study objective: primary objective, secondary objectives, and main contributions; (3) data characteristics: number of slices/scans/cases, imaging modality, organ or anatomical region, pathology or condition, number of centers, dataset source, image characteristics, dataset names, test data, annotations, preprocessing steps, and data quality information; (4) methods: model type, model name, architecture, backbone, pretraining strategy, fine-tuning strategy, model input, task objective, and prompt; (5) evaluation: downstream tasks, evaluation strategy, main performance metrics and scores, and comparisons with the state of the art; (6) implementation details: computing resources and model parameters; (7) FUTURE-AI principles: fairness, universality, traceability, usability, robustness, and explainability; (8) identified gaps and proposed future steps; and (9) community impact: license information, data, model, and code availability.

\subsection{Data items and synthesis of results}

Given the exploratory nature of this scoping review and the marked heterogeneity across datasets, tasks, evaluation protocols, and reporting standards, findings were synthesized descriptively rather than through quantitative meta-analysis.

\subsection{Conceptual framework for evidence mapping}
\label{sec:methods:pillars}

We propose a conceptual framework grounded in three pillars to unify the core characteristics and operational requirements of VFMs in radiology: (i) access to large-scale and diverse imaging data, (ii) scalable pretraining strategies and model architectures, and (iii) adaptation and transferability to downstream clinical tasks. Using this three-pillar framework, we mapped the current evidence to provide a structured characterization of radiology VFMs and to enable consistent comparison across diverse methodological approaches and downstream applications.

\subsubsection{Pillar 1: Large-scale and heterogeneous dataset}

The first pillar captures the scale and composition of the training data used to develop a VFM. Such models require large-scale and clinically heterogeneous imaging datasets encompassing multiple modalities, anatomical regions, pathologies, and patient populations, ideally aggregated across institutions, scanners, and acquisition protocols. Increasing dataset scale and heterogeneity facilitate representation learning, broader domain coverage, and improved generalization capabilities. Table~\ref{tab:data_scalability} summarizes the data-related variables extracted from each included study under this pillar.

\begin{table}[h!]
\centering
\caption{Dataset scale and heterogeneity items.}
\label{tab:data_scalability}
\footnotesize
\setlength{\tabcolsep}{3pt}
\renewcommand{\arraystretch}{1.3}

\begin{tabular}{@{}m{0.28\columnwidth}m{0.68\columnwidth}}
\toprule
\textbf{Item} & \textbf{Description} \\
\toprule
\textbf{\# Data (Unit)} & Number of imaging samples used to train the model categorized into slices or scans.\\
\hline
\textbf{Modality} & Imaging modalities represented in the dataset typically CT, X-ray, US or MRI.\\
\hline
\textbf{Localization} & Anatomical regions included in the dataset. "Multiple" refers to different localizations.\\
\hline
\textbf{Clinical indication} & Clinical conditions or patient populations represented. "Multiple" implies different pathologies.\\
\hline
\textbf{Centers} & Number of contributing data sources or institutions (single center or multicenter) \\
\hline
\textbf{Availability} & Whether the dataset is publicly available or proprietary (public and private) \\
\bottomrule
\end{tabular}
\end{table}

\subsubsection{Pillar 2: Large-scale pretraining and architecture}

The second pillar focuses on scalable pretraining strategies and model architectures. VFMs aim to learn robust, task-agnostic representations that can be efficiently transferred across diverse clinical applications. Consequently, model architectures must accommodate large data volumes and increasing model capacity, capture long-range dependencies, and efficiently process high-resolution and volumetric medical imaging data. Scalable pretraining paradigms therefore play a central role in enabling transferable representations across complex radiological domains. Table~\ref{tab:pill2} summarizes the model and pretraining-related variables extracted from each included study under this pillar. 

\begin{table}[h!]
\centering
\caption{Pretraining and architecture items.}
\label{tab:pill2}
\footnotesize
\setlength{\tabcolsep}{3pt}
\renewcommand{\arraystretch}{1.3}
\begin{tabular}{@{}m{0.32\columnwidth}m{0.64\columnwidth}}
\toprule
\textbf{Item} & \textbf{Description} \\ 
\toprule
\textbf{Model name} & Model’s published name or identifier. \\
\hline
\textbf{Architecture} & Backbone architecture used for the model. \\
\hline
\textbf{Pretraining strategy} & Learning paradigm used. \\
\hline
\textbf{Hardware} & Computational resources for pretraining. \\
\hline
\textbf{Parameters} & Total number of trainable parameters. \\ 
\hline
\textbf{Weights} & Whether pretrained weights are publicly accessible (Yes/No). \\
\hline
\textbf{Code} & Whether training or inference code is publicly available (Yes/No). \\
\bottomrule
\end{tabular}
\end{table}

\subsubsection{Pillar 3: Scalability to multiple applications}

The third pillar characterizes the VFMs' ability to generalize and adapt across multiple downstream clinical tasks while maintaining meaningful performance. This pillar reflects the practical utility, transferability and, robustness of learned representations across diverse radiological applications. Table~\ref{tab:pill3} provides a summary of the downstream evaluation and adaptation variables extracted from each included study under this pillar.

\begin{table}[h!]
\centering
\caption{Downstream evaluation items.}
\label{tab:pill3}
\setlength{\tabcolsep}{3pt}
{\footnotesize
\renewcommand{\arraystretch}{1.3}
\begin{tabular}{@{}m{0.32\columnwidth}m{0.64\columnwidth}}
\toprule
\textbf{Item} & \textbf{Description}  \\
\toprule
\textbf{Tasks} & Number of downstream evaluation tasks performed. \\
\hline
\textbf{Task types} & Categories of tasks addressed (e.g., classification, segmentation). \\
\hline
\textbf{Fine-tuning strategy} & Fine-tuning strategy used, if applicable.\\
\hline
\textbf{Generalization} & Type of generalization shift evaluated in the downstream tasks. \\
\hline
\textbf{Zero-shot (ZS)} & Indicates whether zero-shot evaluation was performed (Yes/No). \\
\hline
\textbf{Few-shot (FS)} & Indicates whether few-shot evaluation was conducted (Yes/No). \\
\hline
\textbf{Baseline outperforming (PT/FM ↑)} & Whether the pretrained or FMs outperformed baseline pretrained models (Yes/Partially/No). \\
\hline
\textbf{Task-specific outperforming (Task-spec. ↑)} & Whether performance exceeded that of task-specific models (Yes/Partially/No) \\
\bottomrule
\end{tabular}
}
\end{table}

\subsection{Future-AI principles alignment}

To complement the technical mapping of VFM studies in radiology, we additionally characterized their reported alignment with principles for trustworthy and deployable artificial intelligence (AI) in healthcare. For this purpose, we adopted the FUTURE-AI framework \citep{futureaiguidelines}, an international consensus guideline developed to support the responsible development, evaluation, and deployment of AI systems across the healthcare lifecycle. The framework defines six core principles—fairness, universality, traceability, usability, robustness, and explainability—and outlines best-practice recommendations spanning technical, clinical, ethical, and regulatory domains.

Alignment with FUTURE-AI principles was descriptively assessed as "reported" or "not reported". A principle was considered reported if the study explicitly described at least one design choice, evaluation procedure, or implementation element corresponding to that principle. No attempt was made to assess completeness, quality or degree of adherence, as the objective was to map reporting patterns.
\section{Results}

\subsection{Study Selection}

\begin{figure}[h]
   \centering
   \includegraphics[width=\textwidth]{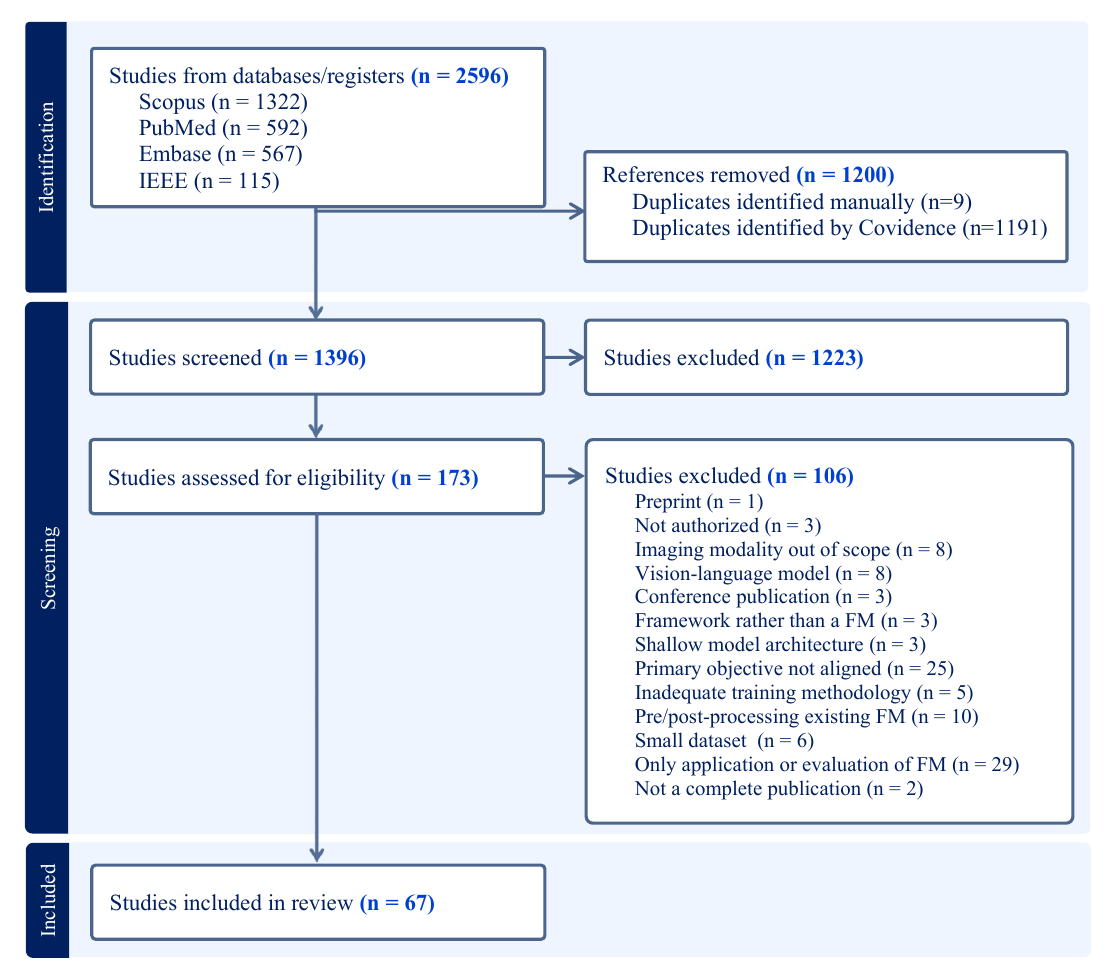}
   \caption{PRISMA Flow Diagram}
   \label{fig:prisma_diagram}
\end{figure}

The study selection process is summarized in the PRISMA flow diagram in Figure~\ref{fig:prisma_diagram}. The database search identified 2,596 records, including 1322 from Scopus, 592 from PubMed, 567 from Embase, and 115 from IEEE. After removal of 1200 duplicate records (1191 identified automatically using Covidence and nine identified manually), 1396 unique studies remained for title and abstract screening. Of these, 1223 studies were excluded, leaving 173 articles for full-text eligibility assessment. Following full-text review, 106 studies were excluded according to the criteria defined in Section~\ref{subsec:eligibility_criteria}. The main reasons for exclusion included preprints, lack of authorization, conference publications, use of non-radiological imaging modalities, incorporation of text-based data, focus on methodological model training frameworks, use of shallow model architectures, and study objectives not aligned with the development of a radiology VFM. Ultimately, 67 studies met the inclusion criteria and were included in the final scoping review. The temporal distribution of included studies shows that eligible studies began to meet the inclusion criteria from late 2023 onward, with a notable shift in trend beginning in 2025, when the number of included studies increased markedly (Figure \ref{fig:pillars}C).

\subsection{Pillar 1: Large-scale and heterogeneous dataset}

Across the 67 included studies, substantial variability was observed in both data dimensionality (2D, 3D and 4D) and dataset scale, with cohort sizes ranging from <100k to >1M samples. For consistency, dataset sizes were reported using units aligned with the underlying data representation: 2D datasets are quantified in slices, whereas 3D and 4D datasets are generally quantified in scans. However, in two studies \citep{bhatt2023, lin2025samct} involving volumetric data, dataset size was reported only in terms of extracted slices, and this convention was therefore retained. An additional category, denoted as ‘multiple’, was introduced for studies combining heterogeneous dimensionalities. 

To facilitate comparison across studies, datasets were further grouped by scale magnitude. For slice-based datasets, small, medium, and large scales corresponded to <100k, 100k–1M, and >1M samples, respectively, whereas for scan-based datasets the corresponding ranges were <10k, 10k–100k, and >100k. Notably, 3D and 4D acquisitions generally encode substantially richer spatial and temporal information than 2D data, even when the reported sample counts are similar.

Among slice-based datasets, six studies (8.9\%) were categorized as small \citep{bhatt2023, alhammuri2025, lin2025perceptguide, chen2025mofo, deng2026, jiang2026ultrasam}, seven (10.3\%) as medium \citep{ma2025, kang2024, zhang2025medsegx, ma2026, yang2025chexfound, yao2026, meyer2026}, and only five (7.4\%) used large-scale datasets \citep{wang2025, lin2025samct, jiao2024, jiang2025ultrafedfm, kang2025}.
Scan-based datasets were predominantly distributed across the small- and medium-scale categories, comprising 15 (22.3\%) \citep{tang2025, pai2024, han2025, silva2025, cui2023, shen2025, zhang2025, yang2025, sun2025, gu2025, machado2025, yang2025brainsn, barbano2026, yang2025crcfound, tang2026} and 17 (25.3\%) \citep{yu2024, yang2024, ding2025, gao2025, wood2024, suo2025, gong2025, cox2024, lei2025, luo2025, chen2024brain, mcconnell2026, li2026, xu2025, tak2026, wang2026fcn, zhang2025bdfm} studies, respectively. Only two studies (3\%) employed large datasets \citep{gao2025lctfound, wu2026}. 
 
Studies categorized as "multiple" were predominantly large-scale when dataset size was reported in slices (7, 10.4\%) \citep{schafer2024, chu2025, gu2024, ma2024, archit2025, zedda2025, jacob2025}. In contrast, three scan-based multimodal datasets (4.4\%) were medium-scale \citep{xie2026, wang2025sammed3d, zhao2026}, with the exception of \citep{he2024}, which was categorized as small.


Modality composition further differentiated studies. Most studies (39, 58.2\%) relied on a single imaging modality \citep{yu2024,tang2025,bhatt2023,pai2024,wang2025,yang2024,han2025,alhammuri2025,silva2025,ding2025,cui2023,lin2025samct,lin2025perceptguide,chen2025mofo,lei2025,ma2025,machado2025,jiao2024,kang2024,mcconnell2026,li2026,yang2025brainsn,tak2026,wang2026fcn,gao2025lctfound,barbano2026,ma2026,zhang2025bdfm,yang2025chexfound,yang2025crcfound,yao2026,jiang2025ultrafedfm,tang2026,wu2026,jacob2025,jiang2026ultrasam,meyer2026,kang2025}. These included 12 studies using single-parametric MRI, 12 CT, nine US, and five X-ray datasets.
In contrast, 26 studies (38.8\%) explicitly incorporated multimodal imaging data \citep{he2024,schafer2024,chu2025,gao2025,wood2024,suo2025,shen2025,gong2025, chen2024masam,zhang2025,cox2024,yang2025,sun2025,gu2025,gu2024,luo2025,chen2024brain,ma2024, zhang2025medsegx,xu2025,deng2026,xie2026,archit2025,zedda2025,wang2025sammed3d,zhao2026}, combining modalities such as CT, single- or multiparametric MRI, PET, US, X-rays, as well as additional medical imaging domains including histopathology, optical coherence tomography, and endoscopy.

Anatomical coverage also varied. Single-location studies (34, 50.7\%) focused on specific regions such as the chest \citep{wang2025,ma2025, yu2024, gao2025lctfound, mcconnell2026, ma2026, yang2025chexfound, yao2026}, colorectum \citep{yang2025crcfound}, heart \citep{jacob2025}, tongue \citep{alhammuri2025}, thyroid \citep{kang2024}, and breast \citep{luo2025}. Among these, the majority (21, 61.7\%) focused on the brain \citep{bhatt2023,yang2024,han2025,he2024,ding2025,
gao2025,wood2024,cui2023,suo2025,zhang2025,cox2024,yang2025,sun2025,chen2024brain,li2026,yang2025brainsn,tak2026,wang2026fcn,barbano2026,zhang2025bdfm,deng2026}. 
In contrast, 30 studies (44.8\%) investigated multiple anatomical locations \citep{tang2025,pai2024,schafer2024,chu2025,silva2025,shen2025,gong2025,chen2024masam,lin2025samct,lin2025perceptguide,chen2025mofo,gu2025,gu2024,lei2025,machado2025,jiao2024,ma2024,zhang2025medsegx,xu2025,xie2026,jiang2025ultrafedfm,tang2026,wu2026,archit2025,zedda2025,wang2025sammed3d,zhao2026,jiang2026ultrasam,meyer2026,kang2025}. The most frequently represented anatomical regions were the brain, thorax, and abdomen, followed by frequent inclusion of liver, kidney, breast, heart, bone, and vascular structures. Several studies also explicitly addressed whole-body or large multi-organ settings.

Pathology coverage was similarly heterogeneous. Multi-pathology studies (22, 32.8\%) encompassed diverse and often pathology-agnostic conditions  \citep{schafer2024,chu2025,silva2025,shen2025,lin2025samct,gu2024,mcconnell2026,zhang2025medsegx,xu2025,zhang2025bdfm,deng2026,xie2026,jiang2025ultrafedfm,tang2026,wu2026,archit2025,zedda2025,wang2025sammed3d,zhao2026,jiang2026ultrasam,meyer2026,kang2025}. Within this group, the most frequently co-occurring disease categories were cancer, neurologic conditions, and thoracic or abdominal diseases, often combined with healthy anatomy. Single-disease studies were distributed across pulmonary/thoracic diseases (7, 10.4\%) \citep{yu2024, wang2025, gao2025lctfound, ma2025, ma2026, yang2025chexfound, yao2026}, neurologic and psychiatric conditions (12, 17.9\%) \citep{yang2024, han2025, ding2025, cui2023, suo2025, zhang2025, cox2024, yang2025, yang2025brainsn,tak2026, wang2026fcn, barbano2026} and cancer-related applications (13, 19.4\%) \citep{bhatt2023, pai2024, gong2025, chen2024masam, lin2025perceptguide, chen2025mofo, machado2025, jiao2024, luo2025, chen2024brain, ma2024, li2026, yang2025crcfound}. A smaller number of studies focused on healthy or non-pathology data (5, 7.4\%) \citep{tang2025, he2024, gao2025, wood2024, lei2025}. Specialized single-disease categories appeared less frequently, such as speech-related conditions \citep{alhammuri2025}, musculoskeletal disorders \citep{gu2025}, pediatric applications \citep{sun2025}, and endocrine disorders in \citep{kang2024}. 

Institutional heterogeneity revealed that most studies used multicenter data (56, 83.5\%) \citep{yu2024, tang2025, wang2025, yang2024, han2025, he2024, schafer2024, chu2025, silva2025, ding2025, cui2023, shen2025, gong2025, chen2024masam, lin2025samct, lin2025perceptguide, zhang2025, cox2024, yang2025, chen2025mofo, sun2025, gu2024, lei2025, ma2025, machado2025, jiao2024, luo2025, ma2024}. However, only seven studies (10.44\%) were single-center \citep{pai2024, alhammuri2025, gao2025, suo2025, gu2025, kang2024, chen2024brain}. Notably, some single-center studies still achieved substantial acquisition diversity, for instance, \citep{gao2025} used data from 14 scanners across five manufacturers, \citep{suo2025} included 14 scanners, and \citep{chen2024brain} also reported 14 scanners within a single institution. 

Regarding accessibility, publicly available datasets were used in 37 studies (55.2\%) \citep{tang2025,pai2024,yang2024,han2025,he2024,schafer2024,chu2025,silva2025,ding2025,cui2023,shen2025,chen2024masam,lin2025samct,lin2025perceptguide,cox2024,yang2025,gu2024,lei2025,ma2025,machado2025,ma2024,li2026,yang2025brainsn,wang2026fcn,barbano2026,ma2026,zhang2025bdfm,deng2026,yang2025chexfound,xie2026,tang2026,wu2026,archit2025,zedda2025,zhao2026,jiang2026ultrasam,meyer2026} and combined with private data in 14 studies (20.9\%) \citep{yu2024,gong2025,zhang2025,chen2025mofo,jiao2024,mcconnell2026,zhang2025medsegx,xu2025,tak2026,yao2026,jiang2025ultrafedfm,wang2025sammed3d,jacob2025,kang2025}, whereas 12 studies (17.9\%) relied exclusively on proprietary datasets \citep{wang2025,alhammuri2025,gao2025,wood2024,suo2025,sun2025,gu2025,kang2024,luo2025,chen2024brain,gao2025lctfound,yang2025crcfound}. 

Table~\ref{tab:pillar_1} and Figure~\ref{fig:pillars} A-G summarize the distribution of dataset scale and heterogeneity across the included studies.

\begin{table}[p]
\centering
\vspace*{-1.3cm}

\makebox[\textwidth][c]{%
\begin{minipage}{\dimexpr\textwidth+2.2cm\relax}
\centering

\caption{Summary of study characteristics under Pillar 1.}
\label{tab:pillar_1}
\setlength{\tabcolsep}{1.2pt}
\renewcommand{\arraystretch}{0.85}
\scriptsize
\rowcolors{2}{gray!3}{gray!8}

\newcolumntype{C}[1]{>{\centering\arraybackslash}p{#1}}

\begin{adjustbox}{
  max width=\linewidth,
  max height=\dimexpr\textheight+1.4cm\relax,
  keepaspectratio
    }
\begin{tabular}{
C{1cm}
C{3.5cm}
C{5cm}
C{1.5cm}
C{4cm}
C{1.75cm}
C{1.5cm}}
\toprule
\textbf{Study} & \textbf{\#Data (Unit)} & \textbf{Modality} & \textbf{Localization} & \textbf{Clinical Indication} & \textbf{Centers} & \textbf{Availability} \\
\midrule

\cite{yu2024} & 10,290 scans & CT & Lungs & Pulmonary/Thoracic Diseases & Multiple & Mixed \\
\cite{tang2025} & 6,814 scans & CT & Multiple & Healthy/No Pathology & Multiple & Public \\
\cite{bhatt2023} & 3,064 slices & MRI (CE-T1) & Brain & Cancer/Tumours & N/R & N/R \\
\cite{pai2024} & 5,513 scans & CT & Multiple & Cancer/Tumours & Single & Public \\
\cite{wang2025} & 1,053,791 slices & X-ray & Thorax & Pulmonary/Thoracic Diseases & Multiple & Private \\
\cite{yang2024} & 64,584 slices & rs-fMRI & Brain & Neurologic/Psychiatric & Multiple & Public \\
\cite{han2025} & 4,409 scans & rs-fMRI & Brain & Neurologic/Psychiatric & Multiple & Public \\
\cite{he2024} & 1,365 scans & sMRI + fMRI & Brain & Healthy/No Pathology & Multiple & Public \\
\cite{schafer2024} & $>$10M slices & CT, MRI, X-ray, WSI, endoscopy, histology, dermoscopy, microscopy. & Multiple & Multiple & Multiple & Public \\
\cite{alhammuri2025} & 50,000 slices & US & Tongue & Speech-related & Single & Private \\
\cite{chu2025} & 17M slices & CT, MRI, X-ray, US, OCT, Retina, Dermoscopy, microscopy & Multiple & Multiple & Multiple & Public \\
\cite{silva2025} & 2,042 scans & CT & Multiple & Multiple & Multiple & Public \\
\cite{ding2025} & 19,687 scans & MRI (T1) & Brain & Neurologic/Psychiatric & Multiple & Public \\
\cite{gao2025} & 57,621 scans & mpMRI (T1, T2, FLAIR, CE-T1) & Brain & Healthy/No Pathology & Single & Private \\
\cite{wood2024} & 64,740 scans & MRI (T1, T2, FLAIR, DWI, SWI) & Brain & Healthy/No Pathology & Multiple & Private \\
\cite{cui2023} & 1,481 scans & rs-fMRI & Brain & Neurologic/Psychiatric & Multiple & Public \\
\cite{suo2025} & 75,861 scans & mpMRI (T1, T2, FLAIR) & Brain & Neurologic/Psychiatric & Single & Private \\
\cite{shen2025} & 1,444 scans & CT, MRI & Multiple & Multiple & Multiple & Public \\
\cite{gong2025} & 40,000 scans & CT, MRI & Multiple & Cancer/Tumours & Multiple & Mixed \\
\cite{chen2024masam} & 932 subjects & CT, MRI, surgical video & Multiple & Cancer/Tumours & Multiple & Public \\
\cite{lin2025samct} & 1.1M slices (5M masks) & CT & Multiple & Multiple & Multiple & Public \\
\cite{lin2025perceptguide} & 33,111 slices & US & Multiple & Cancer/Tumours & Multiple & Public \\
\cite{zhang2025} & 6,585 scans & MRI (T1, CE-T1, T2, FLAIR, DWI) & Brain & Neurologic/Psychiatric & Multiple & Mixed \\
\cite{cox2024} & 82,800 scans & MRI (T1, CE-T1, T2, FLAIR) & Brain & Neurologic/Psychiatric & Multiple & Public \\
\cite{yang2025} & 2,798 scans & T1 MRI; FDG-PET & Brain & Neurologic/Psychiatric & Multiple & Public \\
\cite{zhou2025} & N/A & N/A & N/A & Other/Not Specified & N/A & N/A \\
\cite{chen2025mofo} & 7,039 slices & US & Multiple & Cancer/Tumours & Multiple & Mixed \\
\cite{sun2025} & 516 scans & MRI (T1/T2) & Brain & Pediatric & Multiple & Private \\
\cite{gu2025} & 320 scans & MRI (T1, T2, PD) & Multiple & MSK Conditions & Single & Private \\
\cite{chen2025} & N/A & N/A & N/A & Other/Not Specified & N/A & N/A \\
\cite{gu2024} & 1.35M slices & CT, MRI, US & Multiple & Multiple & Multiple & Public \\
\cite{lei2025} & 14,012 scans & CT & Multiple & Healthy/No Pathology & Multiple & Public \\
\cite{ma2025} & 700,000 slices & X-ray & Thorax & Pulmonary/Thoracic Diseases & Multiple & Public \\
\cite{machado2025} & 2,374 scans & CT & Multiple & Cancer/Tumours & Multiple & Public \\
\cite{jiao2024} & 2,187,915 slices & US & Multiple & Cancer/Tumours & Multiple & Mixed \\
\cite{sun2024} & N/A & N/A & N/A & Other/Not Specified & N/A & N/A \\
\cite{kang2024} & 290,675 slices & US & Thyroid & Endocrine Disorders & Single & Private \\
\cite{luo2025} & 15,660 scans & mpMRI (DCE, T2, DWI) & Breast & Cancer/Tumours & Multiple & Private \\
\cite{chen2024brain} & 57,621 scans & mpMRI(T1, CE-T1, T2, FLAIR) & Brain & Cancer/Tumours & Single & Private \\
\cite{ma2024} & 1,123,310 slices (1,570,263 masks) & CT, MRI, X-ray, dermoscopy, fundus, endoscopy, US, MG, OCT, pathology & Multiple & Cancer/Tumours & Multiple & Public \\
\cite{mcconnell2026} & 98,588 scans & Chest low dose CT & Thorax & Multiple & Multiple & Partial \\
\cite{li2026} & 51,029 scans & MRI & Brain & Cancer/Tumors & Multiple & Public \\
\cite{yang2025brainsn} & 6,629 scans & fMRI & Brain & Neurologic/Psychiatric & Multiple & Public \\
\cite{zhang2025medsegx} & 804,305 slices & CT, MRI, X-ray, US, endoscopy, CTA, CBCT, fundus, dermoscopy & Multiple & Multiple & Multiple & Partial \\
\cite{xu2025} & 98,815 scans & MRI, CT, PET & Multiple & Multiple & Multiple & Partial \\
\cite{tak2026} & 32,015 scans & MRI & Brain & Neurologic/Psychiatric & Multiple & Partial \\
\cite{wang2026fcn} & 10,718 scans & fMRI & Brain & Neurologic/Psychiatric & Multiple & Public \\
\cite{gao2025lctfound} & 105,184 scans & CT & Lung & Pulmonary/Thoracic Diseases & Multiple & Private \\
\cite{barbano2026} & 7,908 scans & MRI (T1) & Brain & Neurologic/Psychiatric & Multiple & Public \\
\cite{ma2026} & 704,363 slices & Chest X-ray & Thorax & Pulmonary/Thoracic Diseases & Multiple & Public \\
\cite{zhang2025bdfm} & 15,300 scans & MRI & Brain & Multiple & Multiple & Public \\
\cite{deng2026} & 4,451 scans & MRI, PET & Brain & Multiple & Multiple & Public \\
\cite{yang2025chexfound} & 987,733 slices & Chest X-ray & Thorax & Pulmonary/Thoracic Diseases & Multiple & Public \\
\cite{yang2025crcfound} & 5,137 scans & CE-CT & Colorectum & Cancer/Tumors & Multiple & Private \\
\cite{xie2026} & 22,000 scans & MRI, CT, X-ray, fundus, microscopy & Multiple & Multiple & Multiple & Public \\
\cite{yao2026} & 520,000 slices & Chest X-ray & Thorax & Pulmonary/Thoracic Diseases & Multiple & Partial \\
\cite{jiang2025ultrafedfm} & 1,015,754 slices & US & Multiple & Multiple & Multiple & Partial \\
\cite{tang2026} & 9,995 scans & CT & Multiple & Multiple & Multiple & Public \\
\cite{wu2026} & 160,000 scans & CT & Multiple & Multiple & Multiple & Public \\
\cite{archit2025} & 3,700,000 slices (15.8M masks) & CT, MRI, endoscopy, US, X-Ray, dermoscopy, ophthalmology, CBCT, fundus, OCT, MG & Multiple & Multiple & Multiple & Public \\
\cite{zedda2025} & 1,350,000 slices & CT, MRI, US & Multiple & Multiple & Multiple & Public \\
\cite{wang2025sammed3d} & 21,729 scans (143518 masks) & CT, MR, US & Multiple & Multiple & Multiple & Partial \\
\cite{zhao2026} & 49k scans (82k masks) & CT, MRI, X-Ray, OCT, US, microscopy, endoscopy, dermoscopy, colonoscopy & Multiple & Multiple & Multiple & Public \\
\cite{jacob2025} & 36,000,000 slices & Cardiac MRI & Heart & Other/Not Specified & Multiple & Partial \\
\cite{jiang2026ultrasam} & 16,820 slices & US & Multiple & Multiple & Multiple & Public \\
\cite{meyer2026} & 186,894 slices (279,765 masks) & US & Multiple & Multiple & Multiple & Public \\
\cite{kang2025} & 1,003,465 slices & US & Multiple & Multiple & Multiple & Partial \\

\bottomrule
\end{tabular}
\end{adjustbox}

\begin{tablenotes}[flushleft]
\scriptsize
\item[] \textit{Abbreviations:}
CBCT, cone beam computed tomography;
CE, contrast enhanced;
CT, computed tomography;
CTA, computed tomography angiography;
DCE, dynamic contrast-enhanced;
DWI, diffusion weighted imaging;
FDG, fluorodeoxyglucose;
FLAIR, fluid-attenuated inversion recovery;
fMRI, functional magnetic resonance imaging;
M, million;
MG, mammography;
mpMRI, multiparametric magnetic resonance imaging;
MRI, magnetic resonance imaging;
N/A, not applicable;
N/R, not reported;
OCT, optical coherence tomography;
PD, proton density;
PET, positron emission tomography;
rs-fMRI, resting-state functional magnetic resonance imaging;
sMRI, structural magnetic resonance imaging;
SWI, susceptibility weighted imaging;
US, ultrasound;
WSI, whole slide imaging.
\end{tablenotes}

\end{minipage}%
}
\vspace*{-0.5cm}

\end{table}

\begin{figure*}[h!]
    \centering
    \includegraphics[width=\textwidth]{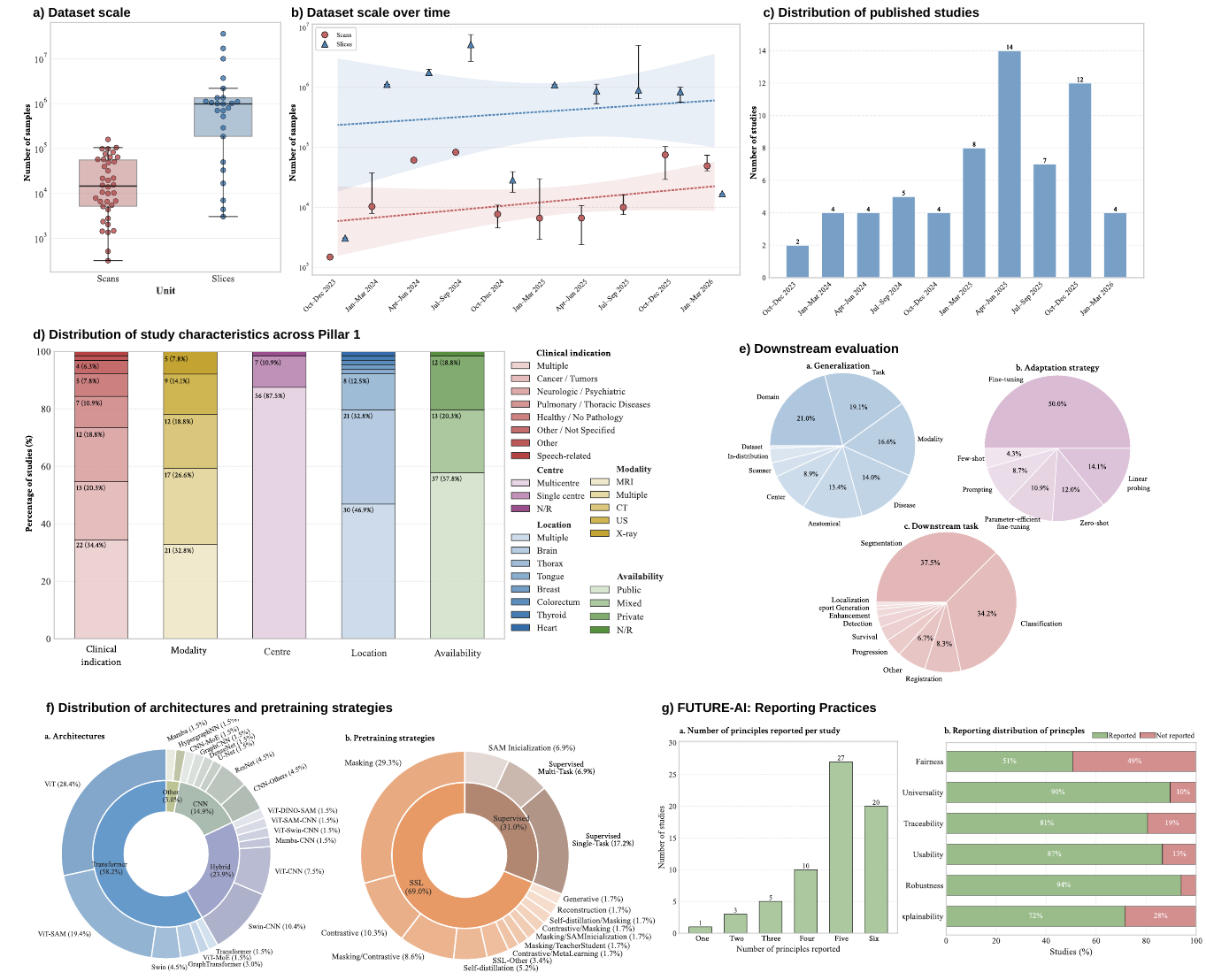}
    \caption{\textbf{Overview of the characteristics of radiology VFMs.} (A) Distribution of dataset scale according to reporting unit. (B) Dataset scale over time: median dataset size and interquartile range, with trend lines and confidence bands. (C) Distribution of studies included in the review. (D) Distribution of study characteristics across Pillar 1. Counts and percentages are displayed within each category block, except for those represented by a single study. (E) Characteristics of downstream evaluation under Pillar 3, including generalization settings (a), adaptation strategies (b), and downstream task categories (c). Category percentages below 3\% are omitted for clarity. (F) Distribution of model architectures (a) and pretraining strategies (b). (G) Reporting of FUTURE-AI principles, including the number of principles reported per study (a) and the proportion of studies addressing each principle (b). 
}
    \label{fig:pillars}
\end{figure*}


\subsection{Pillar 2: Large-scale pretraining and architecture}

Transformer-based models were the most prevalent, appearing in 53 studies (79.1\%) \citep{wang2025,yang2024,he2024,schafer2024,alhammuri2025,chu2025,silva2025,ding2025,gao2025,suo2025,shen2025,chen2024masam,lin2025samct,lin2025perceptguide,cox2024,zhou2025,chen2025mofo,gu2025,chen2025,gu2024,ma2025,machado2025,jiao2024,sun2024,kang2024,chen2024brain,ma2024,mcconnell2026,li2026,yang2025brainsn,zhang2025medsegx,xu2025,tak2026,wang2026fcn,gao2025lctfound,ma2026,zhang2025bdfm,deng2026,yang2025chexfound,yang2025crcfound,xie2026,yao2026,jiang2025ultrafedfm,tang2026,wu2026,archit2025,zedda2025,wang2025sammed3d,zhao2026,jacob2025,jiang2026ultrasam,meyer2026,kang2025}. These models include Vision Transformers (ViT; e.g., ViT-B and ViT-L), Swin Transformers, and other transformer variants. 

Hybrid architectures integrating convolutional and transformer components were reported in 14 studies (20.9\%), reflecting a strategy to jointly capture local, spatial features and long-range contextual dependencies 
\citep{alhammuri2025,chu2025,silva2025,ding2025,suo2025,lin2025perceptguide,cox2024,chen2025mofo,chen2025,sun2024,kang2024,gao2025lctfound,ma2026,wu2026}. Purely convolutional architectures were used in eight studies (11.9\%) \citep{yu2024, bhatt2023, pai2024, wood2024, gong2025, sun2025, lei2025, barbano2026}, including DenseNet, ResNet, and UNet-based architectures. 

Alternative paradigms were less common. State-space (Mamba-based \citep{mamba}) models were reported in two studies \citep{tang2025, yang2025}. Graph-based architectures were limited to three studies \citep{han2025, cui2023, wang2026fcn}. Three studies adopted mixture-of-experts (MoE) architectures, using formulations such as task-specific expert routing \citep{ding2025}, and modality-specific experts with hierarchical or soft gating to handle missing modalities \citep{zhang2025, luo2025}. 

SSL was the most dominant pretraining strategy, employed in 48 studies (71.6\%). Among these, masking–based SSL approaches were most common (19 studies, 40\%) \citep{tang2025,wang2025,chu2025,ding2025,suo2025,gong2025,gu2024,jiao2024,kang2024,mcconnell2026,li2026,yang2025brainsn,wang2026fcn,zhang2025bdfm,yang2025crcfound,jiang2025ultrafedfm,tang2026,zhao2026,kang2025}, followed by contrastive SSL methods (6 studies, 12.5\%) \citep{yu2024, pai2024, alhammuri2025, tak2026, barbano2026, wu2026}. In addition, generative approaches (1 study) \citep{gao2025lctfound} and self-distillation (DINO-based \citep{dino}) approaches (3 studies) \citep{xu2025,zedda2025,jacob2025} were identified. Nine studies (18.8\%) used multi-stage or composite SSL pipelines, combining mask-based and contrastive learning techniques with strategies such as meta-learning or self-distillation frameworks \citep{yang2024,gao2025,cui2023,shen2025,cox2024,yang2025,chen2024brain,yang2025chexfound,yao2026}.

Supervised pretraining was reported in 14 studies (22.4\%) \citep{bhatt2023,schafer2024,silva2025,wood2024,lin2025perceptguide,zhang2025,chen2025mofo,ma2024,ma2025,luo2025,zhang2025medsegx,ma2026,deng2026,wang2025sammed3d,meyer2026}, typically leveraging large, annotated datasets, multi-task objectives, or cyclical training across heterogeneous sources. 

Initialization from large pretrained models was also common, particularly SAM-based \citep{sam} approaches (15 studies, 22.4\%) \citep{shen2025,chen2024masam,lin2025samct,zhou2025,gu2025,chen2025,gu2024,machado2025,zhang2025medsegx,deng2026,xie2026,archit2025,wang2025sammed3d,jiang2026ultrasam,meyer2026}, and were most often adapted through supervised fine-tuning, frequently leveraging PEFT strategies, adapters, or partial backbone freezing. Several studies further incorporated external pretrained encoders, such as CLIP \citep{clip} or ImageNet-pretrained networks, as part of hybrid initialization schemes \citep{he2024, chen2025mofo, luo2025}.

Model parameter counts were reported in 41 studies (61.2\%), ranging from fewer than 10 million parameters to approximately 1.2 billion parameters, with the largest model being \citep{wu2026}, and a median of 214 million. In studies reporting multiple model variants, the largest configuration was considered.

Computational resources were reported in 54 studies (80.6\%). Among these, 21 studies (31.3\%) employed single-GPU training setups, whereas 33 (49.3\%) used multi-GPU configurations, typically ranging from two to eight GPUs and reaching up to 32 and 64 GPUs in some cases, with NVIDIA A100 GPUs being the most commonly used. In contrast, 14 studies (20.9\%) did not specify the computing resources employed.

Regarding openness and reproducibility, code was available in 48 studies (71.6\%), while pretrained weights were publicly available in 35 studies (52.2\%). 

Architectural and pretraining characteristics of the included studies are summarized in Table~\ref{tab:pillar_2} and represented in Figure~\ref{fig:pillars} F.

\begin{table}[p]
\centering
\vspace*{-1.3cm}

\makebox[\textwidth][c]{%
\begin{minipage}{\dimexpr\textwidth+2.2cm\relax}
\centering

\caption{Summary of study characteristics under Pillar 2.}
\label{tab:pillar_2}
\setlength{\tabcolsep}{1.2pt}
\renewcommand{\arraystretch}{0.83}
\scriptsize
\rowcolors{2}{gray!3}{gray!8}

\newcolumntype{C}[1]{>{\centering\arraybackslash}p{#1}}

\begin{adjustbox}{
  max width=\linewidth,
  max height=\dimexpr\textheight+1.4cm\relax,
  keepaspectratio
    }
\begin{tabular}{
C{1.25cm}
C{2.75cm}
C{3.5cm}
C{3.85cm}
C{2.45cm}
C{2.25cm}
C{1.25cm}
C{1cm}}
\toprule
\textbf{Study} & \textbf{Model name} & \textbf{Architecture} & \textbf{Pretraining strategy} & \textbf{Hardware} & \textbf{Parameters} & \textbf{Weights} & \textbf{Code} \\
\midrule

\cite{yu2024} & DrasCLR & 3D CNN & MoCo Contrastive SSL & 4×V100 & N/R & Yes & Yes \\
\cite{tang2025} & MambaMIM & CNN–Mamba hybrid & MIM & 1×A800 & N/R & Yes & Yes \\
\cite{bhatt2023} & None & U-Net & Supervised & Vertex AI & N/R & No & No \\
\cite{pai2024} & None & 3D ResNet50 & SimCLR & 2×RTX 8000 & 200M & Yes & Yes \\
\cite{wang2025} & None & ViT-L & MAE & 8×A800 & 348M & No & Yes \\
\cite{yang2024} & BrainMass & Transformer & MAE + contrastive & 64×V100 & 67M & Yes & Yes \\
\cite{han2025} & HGFM & Hypergraph & SSL (link prediction) & N/R & N/R & No & No \\
\cite{he2024} & FM-APP & ViT + BiomedCLIP & Masked regressor & N/R & 80–100M & N/R & Yes \\
\cite{schafer2024} & UMedPT & Swin Transformer + decoders & Supervised multitask & N/R & N/R & Yes & Yes \\
\cite{alhammuri2025} & TongueTransUNet & UNet + ViT & SimCLR & Vertex/Colab & 130–180M & No & No \\
\cite{chu2025} & Frepa & ViT/Swin/ConvNeXt & MAE-style SSL & 2×A100 & 32–87M & Yes & Yes \\
\cite{silva2025} & None & Swin-UNETR & Supervised & 4×A6000 & 62–65M & Yes & Yes \\
\cite{ding2025} & DenseFormerMoE & DenseNet + ViT & MAE & 1×RTX 4090 & 22–86M & No & No \\
\cite{gao2025} & None & ViT & MAE + contrastive & 8×A100 & 500–800M & Yes & Yes \\
\cite{wood2024} & None & 3DDenseNet201 & Supervised & 2×RTX 2080 & N/R & Yes & Yes \\
\cite{cui2023} & MeTSK & ST-GCN & Contrastive SSL + meta-learning & N/R & $<$10M & No & No \\
\cite{suo2025} & SwinClassifier & Swin-UNETR & MAE & 8×A100 & N/R & Yes & Yes \\
\cite{shen2025} & ProtoSAM-3D & SAM-Med3D & MAE + distillation & 1×A100 & 10–113M & No & No \\
\cite{gong2025} & ViNet & 3D ResNet-18 & SSL (image restoration) & 1×V100 & N/R & No & No \\
\cite{chen2024masam} & MA-SAM & SAM + 3D adapters & SAM initialization & 8×A100 & $\sim$97M & Yes & Yes \\
\cite{lin2025samct} & SAMCT & SAM + U-Net & SAM initialization & HPC HUST & 120M & Yes & Yes \\
\cite{lin2025perceptguide} & PerceptGuide & Swin-UNet & Supervised multi-task & 4×A4000 & 66M & Yes & Yes \\
\cite{zhang2025} & MoME/MoME+ & nnUNet MoE & Supervised & 1×A100 & N/R & Yes & Yes \\
\cite{cox2024} & BrainSegFounder & 3DSwinUNETR & Dual-stage SSL (MAE, contrastive, rotation prediction) & 64×A100 & 62–69M & Yes & Yes \\
\cite{yang2025} & ADFound & Vim (Mamba) & MAE + contrastive & 1×RTX 4090 & N/R & No & Yes \\
\cite{zhou2025} & MASG-SAM & SAM-ViT-B & SAM init & 1×RTX 4090 & $\sim$98M & No & Yes \\
\cite{chen2025mofo} & MOFO & CSWin Transformer + CNN & ImageNet + CLIP & 4×RTX 3090 & N/R & Yes & Yes \\
\cite{sun2025} & BME-X & DU-Net & No pretraining & N/R & N/R & No & No \\
\cite{gu2025} & SegmentAnyBone & SAM-based & SAM initialization & 1×RTX A6000 & N/R & Yes & Yes \\
\cite{chen2025} & cineCMR-SAM & U-Net + SAM blocks & SAM initialization & DGX-A100 & 1099M & No & Yes \\
\cite{gu2024} & LeSAM & SAM-ViT-B & SAM inicialization + MAE & 1×RTX 4090 & N/R & No & No \\
\cite{lei2025} & MedLAM/MedLSAM & CNN & SSL & 4×RTX 3090 Ti & N/R & Yes & Yes \\
\cite{ma2025} & Ark+ & Swin Transformer-L & Supervised multi-dataset & 4×A100 & $\sim$200M & Yes & Yes \\
\cite{machado2025} & ONCOPILOT & SAM-ViT-B & SAM initialization & 32×V100 & N/R & No & No \\
\cite{jiao2024} & USFM & ViT-B & MIM & 1×A100 & N/R & Yes & Yes \\
\cite{sun2024} & SAM-AutoMed & SAM-Med2D + MobileNet & SAM initialization & 1×RTX 2080 & $\sim$108–110M & No & No \\
\cite{kang2024} & Deblurring MIM & ConViT-B & MIM & N/R & $\sim$90M & Yes & Yes \\
\cite{luo2025} & MOME & BEiT3 + MOME & Adapter training & 1×RTX 3090 & 276M & Yes & Yes \\
\cite{chen2024brain} & ViT AE & 3DViT & MAE + contrastive & 8×A100 & N/R & Yes & Yes \\
\cite{ma2024} & MedSAM & SAM-ViT-B & SAM initialization + supervised & 20×A100 & 93.7M & Yes & Yes \\
\cite{mcconnell2026} & TANGERINE & 3DViT & MAE & 4×A6000 & $\sim$312M & Yes & Yes \\
\cite{li2026} & UMBIF & ViT & ImageNet initialization + MAE & HPC & N/R & No & No \\
\cite{yang2025brainsn} & BrainSN & Transformer & SSL reconstruction & 1×RTX 4090D & N/R & No & Yes \\
\cite{zhang2025medsegx} & MedSegX & SAM-based + MoE & Supervised & N/R & 94/312/641M & No & No \\
\cite{xu2025} & 3DINO-ViT & 3D DINOv2 & DINOv2 & 4×A100-SXM4 & 307M & Yes & Yes \\
\cite{tak2026} & BrainIAC & 3D ViT & SimCLR & N/R & N/R & No & No \\
\cite{wang2026fcn} & None & Graph transformer & SSL reconstruction & N/R & N/R & No & No \\
\cite{gao2025lctfound} & LCTfound & Cross-attention UNet & DDPM SSL & 16×V100 & 200M & Yes & No \\
\cite{barbano2026} & AnatCL & 3DResNet & Contrastive SSL & 1×V100 + 1×A100 & 34M & Yes & Yes \\
\cite{ma2026} & Ark+ & Swin Transformer-CNN & Supervised cyclic & 4×V100 + 4×A100 & N/R & Yes & Yes \\
\cite{zhang2025bdfm} & BDFM & Swin Transformer & MIM & 1×V100 & 91.66/30.95M & No & Yes \\
\cite{deng2026} & SAM-Brain3D+HyDA & 3D SAM-based & SAM-Med3D initialization & 1×A100 & 100.51M & No & Yes \\
\cite{yang2025chexfound} & CheXFound & ViT-L & MAE + distillation & 8×A100 & 307M & No & Yes \\
\cite{yang2025crcfound} & CRCFound & 3DViT & MAE & 4×A100 & N/R & Yes & No \\
\cite{xie2026} & EICSeg & Dual encoder & DINOv2 + SAM inizialization & 8×V100 & 38.8M & No & Yes \\
\cite{yao2026} & EVA-X & Dual ViT & Contrastive + MIM & N/R & 6/22/86M & Yes & Yes \\
\cite{jiang2025ultrafedfm} & UltraFedFM & ViT & Federated MAE & N/R & 86/307/632M & No & No \\
\cite{tang2026} & Hi-End-MAE & ViT + Hierarchical Dense Decoder & MIM & N/R & N/R & No & Yes \\
\cite{wu2026} & VoCo & SwinUNETR & Omni-supervised & 8×H800 & 31M–1.2B & Yes & Yes \\
\cite{archit2025} & MedicoSAM & SAM + ViT-B & SAM initialization & 8×A100 & 86M & Yes & Yes \\
\cite{zedda2025} & Radio DINO & ViT & DINO/DINOv2 & 2×A100 & 5.8M–86M & No & Yes \\
\cite{wang2025sammed3d} & SAM-Med3D & 3DViT & Supervised & 2×A100 & N/R & Yes & Yes \\
\cite{zhao2026} & SegMIC & ViT & MIM & 1×A100 & 114.6M & No & Yes \\
\cite{jacob2025} & None & ViT-S & DINO & 8×H100 & 21M & No & No \\
\cite{jiang2026ultrasam} & UltraSAM & SAM-based & SAM initialization & N/R & 86M & No & No \\
\cite{meyer2026} & UltraSam & SAM-ViT-B & SAM initialization + supervised & 4×H100; 1×A100 & 86M & Yes & Yes \\
\cite{kang2025} & URFM & ViT & MIM & N/R & 86M & Yes & Yes \\

\bottomrule
\end{tabular}
\end{adjustbox}

\begin{tablenotes}[flushleft]
\scriptsize
\item[] \textit{Abbreviations:}
3D, three-dimensional;
AI, artificial intelligence;
B, base;
CLIP, Contrastive Language-Image Pre-training;
CNN, convolutional neural network;
CSWin, cross-shaped window;
DDPM, denoising diffusion probabilistic models;
DINO, distillation with no labels;
DUNet, double UNet;
L, large;
M, million;
MAE, masked autoencoder;
MIM, masked image modeling;
MoCo, momentum contrast;
MoE, mixture of experts;
MOME, mixture of modality experts;
N/R, not reported;
S, small;
SAM, Segment Anything Model;
SimCLR, Simple Framework for Contrastive Learning of Visual Representations;
SSL, self-supervised learning;
ST-GCN, spatial temporal graph convolutional network;
ViT, Vision Transformer;
Vim, vision Mamba.
\end{tablenotes}

\end{minipage}%
}
\vspace*{-0.5cm}

\end{table}


\subsection{Pillar 3: Scalability to multiple applications}

Substantial variability was observed in scalability across tasks, datasets, and clinical settings (Table~\ref{tab:pillar_3}). Most studies (58, 86.5\%) evaluated multiple downstream tasks, ranging from 1 to 146 tasks (median: 6). Notably, 19 studies (28.4\%) evaluated 10 or more tasks \citep{yang2024,he2024,schafer2024,chu2025,silva2025,lin2025samct,lin2025perceptguide,zhang2025,gu2024,jiao2024,ma2024,mcconnell2026,zhang2025medsegx,barbano2026,yao2026,jiang2025ultrafedfm,wu2026,wang2025sammed3d,zhao2026}. 

Common downstream tasks included classification, segmentation, detection, regression, survival analysis, enhancement, localization, and progression modeling. Segmentation (67.1\%) and classification (61.1\%) were the most frequently investigated tasks. 

Fine-tuning (full or partial) was the dominant adaptation strategy (45 studies, 67.1\%),  often complemented by PEFT methods (10 studies, 14.9\%) \citep{silva2025, chen2024masam, lin2025samct, gu2025, chen2025, gu2024, deng2026, xie2026, yang2025chexfound, jiang2026ultrasam}. In contrast, frozen-encoder evaluations using linear probing (LP) were explicitly performed in 12 studies (17.9\%)  \citep{yu2024,pai2024,yang2024,chu2025,ma2025,sun2024,xu2025,tak2026,gao2025lctfound,barbano2026,ma2026,yang2025chexfound}. Similarly, ZS evaluation without task-specific fine-tuning was reported in 12 studies (17.9\%) \citep{he2024,silva2025,cui2023,shen2025,lin2025perceptguide,chen2025mofo,lei2025,ma2025,ma2024,yang2025brainsn,zhang2025medsegx,zhao2026}. FS learning scenarios were evaluated in 17 studies (25.3\%), commonly involving 1–5 labeled samples per task \citep{yang2024,han2025,schafer2024,silva2025,shen2025,lin2025perceptguide,zhou2025,lei2025,ma2025,tak2026,gao2025lctfound,xie2026,yao2026,tang2026,zhao2026,jacob2025,jiang2026ultrasam}.

Generalization of VFMs was evaluated in 97\% of studies. Disease shift was the most frequently evaluated setting (54 studies, 80.5\%) \citep{yu2024,tang2025,pai2024,wang2025,yang2024,han2025,schafer2024,chu2025,ding2025,gao2025,cui2023,suo2025,chen2024masam,lin2025samct,lin2025perceptguide,zhang2025,cox2024,yang2025,zhou2025,chen2025mofo,sun2025,gu2024,ma2025,machado2025,jiao2024,sun2024,kang2024,luo2025,chen2024brain,ma2024,mcconnell2026,li2026,yang2025brainsn,zhang2025medsegx,xu2025,tak2026,wang2026fcn,gao2025lctfound,barbano2026,ma2026,zhang2025bdfm,deng2026,yang2025chexfound,yang2025crcfound,xie2026,yao2026,jiang2025ultrafedfm,zedda2025,wang2025sammed3d,zhao2026,jacob2025,jiang2026ultrasam,meyer2026,kang2025}. Task shift generalization was assessed in 32 studies (47.7\%) \citep{yu2024,pai2024,wang2025,schafer2024,chu2025,ding2025,lin2025perceptguide,sun2025,lei2025,ma2025,machado2025,jiao2024,sun2024,kang2024,luo2025,yang2025brainsn,xu2025,tak2026,wang2026fcn,gao2025lctfound,barbano2026,ma2026,zhang2025bdfm,deng2026,yang2025chexfound,yang2025crcfound,yao2026,jiang2025ultrafedfm,wu2026,zedda2025,jacob2025,meyer2026}. Anatomical shift generalization was reported in 30 studies (44.7\%) \citep{tang2025,pai2024,schafer2024,chu2025,silva2025,shen2025,chen2024masam,lin2025samct,lin2025perceptguide,zhou2025,chen2025mofo,gu2025,gu2024,lei2025,machado2025,jiao2024,ma2024,zhang2025medsegx,xu2025,yang2025chexfound,xie2026,jiang2025ultrafedfm,tang2026,archit2025,zedda2025,wang2025sammed3d,zhao2026,jiang2026ultrasam,meyer2026,kang2025}. Modality shift generalization was evaluated in 19 studies (28.4\%) \citep{tang2025,he2024,wood2024,shen2025,gong2025,chen2024masam,zhang2025,cox2024,yang2025,zhou2025,gu2024,ma2024,zhang2025medsegx,xie2026,tang2026,archit2025,wang2025sammed3d,zhao2026,jiang2026ultrasam}.  In addition, cross-center generalization was explicitly evaluated in 23 studies (34.3\%) \citep{pai2024,wang2025,yang2024,schafer2024,wood2024,gong2025,chen2024masam,chen2025mofo,chen2025,ma2025,machado2025,jiao2024,luo2025,mcconnell2026,li2026,zhang2025medsegx,gao2025lctfound,yang2025crcfound,yao2026,jiang2025ultrafedfm,wang2025sammed3d,jacob2025,jiang2026ultrasam}, while cross-scanner robustness was reported in four studies \citep{jiao2024,luo2025,chen2024brain,jiang2026ultrasam}. Only two studies did not evaluate generalization \citep{alhammuri2025, bhatt2023}. 

Performance evaluation against established pretrained and VFMs was a central component of the included studies, reported in 49 (73.1\%) of cases. Among these, 93.8\% demonstrated superior performance, either consistently across benchmarks or in more than 90\% of the evaluated downstream tasks, while an additional 6\% reported partial improvements, demonstrating superior performance in at least one evaluation setting. Comparisons with fully supervised, task-specific models were even more prevalent, reported in 57 studies (85.1\%). Among these, 84\% outperformed task-specific models, whereas 15.8\% demonstrated partial improvements. 

Downstream evaluation characteristics of the included studies are summarized in Table~\ref{tab:pillar_3} and represented in Figure~\ref{fig:pillars} E.


\begin{table}[p]
\centering
\vspace*{-1.3cm}

\makebox[\textwidth][c]{%
\begin{minipage}{\dimexpr\textwidth+2.2cm\relax}
\centering

\caption{Summary of study characteristics under Pillar 3.}
\label{tab:pillar_3}
\setlength{\tabcolsep}{1.2pt}
\renewcommand{\arraystretch}{0.8}
\scriptsize
\rowcolors{2}{gray!3}{gray!8}

\newcolumntype{C}[1]{>{\centering\arraybackslash}p{#1}}

\begin{adjustbox}{
  max width=\linewidth,
  max height=\dimexpr\textheight+1.4cm\relax,
  keepaspectratio
    }
\begin{tabular}{
C{1.0cm}
C{1.1cm}
C{2.6cm}
C{3.1cm}
C{4.6cm}
C{1.1cm}
C{1.1cm}
C{1.5cm}
C{1.8cm}}
\toprule
\textbf{Study} & \textbf{\#Tasks} & \textbf{Task types} & \textbf{Fine-tuning strategy} & \textbf{Generalization} & \textbf{ZS} & \textbf{FS} & \textbf{PT/FM $\uparrow$} & \textbf{Task-spec.$\uparrow$} \\
\midrule

\cite{yu2024} & 9 & cls, reg, surv, seg & LP, FT & Disease, Task & No & No & Yes & Yes \\
\cite{tang2025} & 8 & seg & FT & Modality, Anatomical & No & No & Yes & Yes \\
\cite{bhatt2023} & 1 & seg & N/R & Task & N/R & N/R & N/R & N/R \\
\cite{pai2024} & 3 & cls, surv & LP, FT & Disease & No & No & Yes & Yes \\
\cite{wang2025} & 3 & cls, rp & FT with multimodal decoder & Task & No & No & Partially & Partially \\
\cite{yang2024} & 15 & cls & LP, FS & Disease & Yes & Yes & Yes & Yes \\
\cite{han2025} & 4 & cls & FS, FT & Disease & No & Yes & N/R & Yes \\
\cite{he2024} & 92 & reg & ZS & Modality, Task & Yes & No & N/R & Yes \\
\cite{schafer2024} & 20 & seg, cls, det & FT, FS & Domain, Modality & No & Yes & Yes & Yes \\
\cite{alhammuri2025} & 1 & seg & Semi-supervised & In-distribution & No & No & N/R & Yes \\
\cite{chu2025} & 32 & seg, cls, det & LP & Modality, Task & Yes & N/R & Yes & Yes \\
\cite{silva2025} & 13 & seg & ZS, FS, FT, PEFT & Anatomical & Yes & Yes & Yes & Yes \\
\cite{ding2025} & 3 & cls, reg & Multi-Task FT & Task & No & No & N/R & Yes \\
\cite{gao2025} & 1 & cls & FT & In-distribution & No & No & N/R & Yes \\
\cite{wood2024} & 1 & reg & Transfer learning & Domain & No & No & N/R & N/R \\
\cite{cui2023} & 1 & cls & ZS & Disease & Yes & No & Yes & Yes \\
\cite{suo2025} & 1 & cls & FT & Dataset shift & No & No & N/R & Yes \\
\cite{shen2025} & 5 & seg & ZS, distillation + FT, FS & Modality, Anatomical, Task & Yes & Yes & Yes & Yes \\
\cite{gong2025} & 1 & cls & FT + experiential guidance & Center, Domain & No & No & N/R & Yes \\
\cite{chen2024masam} & 5 & seg & PEFT, FT & Modality, Task & No & No & Yes & Yes \\
\cite{lin2025samct} & 118 & seg & PEFT & Anatomical & No & No & Yes & Yes \\
\cite{lin2025perceptguide} & 20 & seg, cls & ZS, FT, FS & Anatomical, Task & Yes & Yes & Yes & Yes \\
\cite{zhang2025} & 17 & seg & FT & Modality, Disease & No & No & Yes & Yes \\
\cite{cox2024} & 2 & seg & FT & Disease, Task & No & No & N/R & Yes \\
\cite{yang2025} & 6 & cls, prog & FT & Modality, Disease & No & No & Yes & Yes \\
\cite{zhou2025} & 6 & seg & FS & Task & No & Yes & Yes & Yes \\
\cite{chen2025mofo} & 3 & seg & FT, ZS & Anatomical & Yes & No & Yes & Yes \\
\cite{sun2025} & 4 & enh, seg, reg, cls & N/A & Domain, Scanner & No & No & N/R & Yes \\
\cite{gu2025} & 1 & seg & PEFT & Anatomical, Modality & Yes & No & N/R & Yes \\
\cite{chen2025} & 3 & seg & PEFT & Center, Scanner & No & No & N/R & Yes \\
\cite{gu2024} & 12 & seg & PEFT & Modality, Disease & No & No & Yes & Yes \\
\cite{lei2025} & 3 & loc, seg & ZS, FS & Anatomical & Yes & Yes & Partially & Yes \\
\cite{ma2025} & 8 & cls & FT, LP, FS, ZS & Disease, Center & Yes & Yes & Yes & Yes \\
\cite{machado2025} & 1 & seg & FT & Center & No & No & Yes & Yes \\
\cite{jiao2024} & 11 & seg, cls, enh & FT & Anatomical, Task & No & No & Yes & Yes \\
\cite{sun2024} & 6 & seg, cls & LP & Modality, Task & No & No & Yes & Yes \\
\cite{kang2024} & 4 & cls, seg & FT & Domain, Scanner & No & No & Yes & Yes \\
\cite{luo2025} & 3 & cls & FT & Domain, Center & No & No & N/R & Yes \\
\cite{chen2024brain} & 3 & det, cls & FT & Task & No & No & N/R & Yes \\
\cite{ma2024} & 146 & seg & ZS & Modality, Anatomical, Task & Yes & No & Yes & Yes \\
\cite{mcconnell2026} & 14 & cls & FT & Disease, Center, Domain, In-distribution & No & No & Yes & Partially \\
\cite{li2026} & 2 & cls & FT & Disease, Center, Domain & No & No & Yes & Yes \\
\cite{yang2025brainsn} & 5 & cls, reg, other & ZS, FT & Task, Domain & Yes & No & Yes & Partially \\
\cite{zhang2025medsegx} & 111 & seg & ZS, FT & Disease, Task, Modality, Anatomical, Domain, In-distribution, Center & Yes & No & Yes & Yes \\
\cite{xu2025} & 6 & cls, reg, seg & LP, FT & Modality, Anatomical, Domain & No & No & Yes & N/R \\
\cite{tak2026} & 7 & cls, reg, seg, sup & FT, LP, FS & Disease, Task, Domain & No & Yes & Yes & N/R \\
\cite{wang2026fcn} & 4 & cls, reg, other & FT & Task, Domain & No & No & N/R & Yes \\
\cite{gao2025lctfound} & 8 & cls, seg, prog, other & FT, LP, FS & Disease, Task, Center, Domain & No & Yes & Yes & Yes \\
\cite{barbano2026} & 22 & cls, reg & LP & Disease, Task, Domain & No & No & Yes & Partially \\
\cite{ma2026} & 3 & cls, seg, other & FT, LP & Task, Modality, Domain & No & No & Yes & Yes \\
\cite{zhang2025bdfm} & 3 & cls, seg & FT & Disease, Task, Domain & No & No & N/R & Yes \\
\cite{deng2026} & 3 & cls, seg, prog & PEFT & Disease, Modality, Domain & No & No & Yes & Yes \\
\cite{yang2025chexfound} & 9 & cls, seg, prog & LP, decoder tuning & Disease, Task, Domain, In-distribution & No & No & Yes & N/R \\
\cite{yang2025crcfound} & 8 & cls, sup, prog & PEFT & Task, Center, Domain & No & No & N/R & Yes \\
\cite{xie2026} & 9 & seg & PEFT, FS & Anatomical, Modality, Domain & No & Yes & Yes & Partially \\
\cite{yao2026} & 11 & cls, seg, other & FT, FS & Disease, Task, Center, Domain & No & Yes & Yes & Yes \\
\cite{jiang2025ultrafedfm} & 13 & cls, seg & FT & Disease, Anatomical, Center, Modality, Domain & No & No & Yes & Yes \\
\cite{tang2026} & 9 & seg & FT, FS & Modality, Anatomical, Domain & No & Yes & Yes & N/R \\
\cite{wu2026} & 51 & cls, seg, other & FT & Task, Anatomical, Domain & No & No & Yes & N/R \\
\cite{archit2025} & 4 & seg & FT & Task, Modality, Domain & No & No & Yes & Partially \\
\cite{zedda2025} & 7 & cls, seg & FT & Modality, Task, Domain & No & No & Yes & Partially \\
\cite{wang2025sammed3d} & 16 & seg & FT & Disease, Anatomical, Modality, Domain & No & No & Yes & Partially \\
\cite{zhao2026} & 56 & seg & ZS, FS & Anatomical, Modality, Domain & Yes & Yes & Yes & N/R \\
\cite{jacob2025} & 9 & cls, seg, other & FT, FS & Task, Modality, Center, Scanner, Domain & No & Yes & Partially & Partially \\
\cite{jiang2026ultrasam} & 10 & seg & PEFT, FS & Anatomical, Center, Scanner, Modality, Domain & Yes & Yes & Yes & Yes \\
\cite{meyer2026} & 3 & cls, seg, other & FT & Anatomical, Modality, Domain & No & No & Yes & N/R \\
\cite{kang2025} & 10 & cls & FT & Anatomical, Domain & No & No & Yes & N/R \\

\bottomrule
\end{tabular}
\end{adjustbox}

\begin{tablenotes}[flushleft]
\scriptsize
\item[] \textit{Abbreviations:}
cls, classification;
det, detection;
enh, image enhancement;
FS, few-shot;
FT, fine-tuning;
loc, localization;
LP, linear probing;
N/A, not applicable;
N/R, not reported;
PEFT, parameter-efficient fine-tuning;
prog, prognosis;
PT/FM $\uparrow$, superior performance over pretrained or foundation models;
reg, image registration;
rp, report generation;
surv, survival;
Task-spec.$\uparrow$, superior performance over task-specific models;
ZS, zero-shot.
\end{tablenotes}

\end{minipage}%
}
\vspace*{-0.5cm}

\end{table}

\subsection{FUTURE-AI principles alignment}
The alignment of the included studies with the FUTURE-AI principles is summarized in Figure~\ref{fig:pillars} G. To enhance transparency and reproducibility, a detailed breakdown is provided in Table 8 (Supplementary Material,  Section 7.3), where each reported principle is accompanied by a brief justification outlining the specific criteria and evidence supporting its classification.   

Across the included literature, adherence to the FUTURE-AI principles was heterogeneous and inconsistently reported. Only 20 studies (29.9\%) comprehensively addressed all principles  \citep{yang2024, chu2025, cui2023, gong2025, cox2024, chen2025mofo, lei2025, luo2025, ma2024, mcconnell2026, li2026, yang2025brainsn, tak2026, barbano2026, ma2026, yao2026, jiang2025ultrafedfm, zedda2025, jacob2025, kang2025}. 

Among individual principles, robustness was the most frequently reported (63 studies, 94\%), reflecting the strong emphasis on performance stability and generalization across datasets and tasks. Universality (60 studies, 89.6\%) and usability (58 studies, 86.6\%) were also commonly addressed, consistent with the focus on scalability and practical applicability of VFMs across diverse clinical settings. Traceability was reported in 54 studies (80.5\%), typically in the form of methodological transparency or documentation of model development and evaluation. In contrast, explainability and fairness were less consistently evaluated, reported in 48 (71.6\%) and 34 (50.7\%) studies, respectively.

\section{Discussion}
This scoping review aims to provide a comprehensive overview of the current landscape of VFMs in radiology. Although the topic has gained substantial visibility in recent years, 
the scientific evidence remains recent and rapidly evolving. The first studies meeting the inclusion criteria emerged only in late 2023, with publication activity increasing markedly from 2025 onward (Figure~\ref{fig:pillars}). This growth  likely reflects the convergence of several technological and methodological advances, including the success of FMs in computer vision, the maturation of SSL approaches, increased access to large-scale imaging datasets, and progress in scalable computational infrastructures \citep{CHEN2026100123}.

In total, 67 studies met the predefined eligibility criteria. While this number may appear modest relative to the growing interest in the field, it should be interpreted in the context of the deliberately focused scope of this review. Specifically, we restricted our analysis to imaging-only FM and excluded approaches incorporating language data. This decision was made to maximize methodological comparability across studies and evaluation settings while isolating imaging-specific contributions from the additional complexities associated with multimodal learning and textual integration \citep{cheng2025understanding}. As a result, this review provides a focused assessment of the state of VFM in radiology, enabling the identification of prevailing trends, methodological limitations, and priorities for future research.

Among the most important observations arising from this review is the absence of a consistently applied definition of VFMs in radiology. This lack of consensus is reflected in the substantial heterogeneity observed across studies, encompassing dataset scale and composition, pretraining paradigms, architectural design choices, fine-tuning approaches, and downstream evaluation tasks. To provide a more structured analysis, we organized the literature according to three core pillars: data scale and heterogeneity, architectural and pretraining scalability, and downstream transferability and generalization. 

Regarding the first pillar, current radiology VFMs remain strongly shaped by the scale of the datasets used during pretraining. Although slice-based datasets occasionally reach the million-sample scale, scan-based datasets remain predominantly small to medium in size. This discrepancy reflects intrinsic barriers in radiology, including the cost of data acquisition, storage, and annotation, as well as privacy constraints. Importantly, it raises questions regarding the extent to which current models can genuinely be considered “foundation models”, as their scale often falls short of what is typically observed in natural image or multimodal FM development \citep{de2025foundation}. 

Dataset composition across imaging modalities, anatomical regions, and pathologies also revealed important observations. In principle, datasets encompassing multiple imaging modalities should benefit from learning complementary information, enabling VFMs to develop more robust and transferable visual representations \citep{moor2023foundation}. Despite these potential benefits, more than half of the reviewed studies relied on a single imaging modality. This under-representation was particularly evident for nuclear medicine modalities, such as PET, which appeared in only a small fraction of studies. One potential factor limiting the integration of heterogeneous datasets across imaging modalities is the substantial differences in spatial dimensionality. These differences often encourage the use of slice-wise or pseudo-3D processing strategies, even for inherently volumetric modalities such as CT and MRI, thereby limiting the exploitation of three-dimensional contextual information\citep{noh2025narrative}. Consequently, explicit volumetric representation learning was supported by relatively few studies. Temporal modeling was even less frequently addressed, likely due to the limited availability of longitudinal (4D) imaging datasets, which constraints the ability of current VFMs to capture disease evolution over time. 

Pathology coverage was similarly imbalanced, with some studies using multi-pathology datasets to support broader representation learning, whereas many remained focused on single-disease settings, most commonly involving thoracoabdominal CT in oncology, brain MRI in neurological disorders, and chest X-ray imaging in pulmonary diseases. In contrast, several major radiological domains, such as cardiovascular, pediatric, musculoskeletal, obstetric, and inflammatory imaging, remained absent or sparsely represented. Although specialized datasets often align with well-defined clinical tasks, they may not fully exploit the potential of FMs to learn transferable medical representations. The coexistence of these two paradigms suggests that the field has not yet converged on a unified strategy for balancing task-specific performance with general-purpose modeling.

Data heterogeneity, accessibility, and reproducibility also pose significant challenges for VFMs. While the predominance of multicenter studies suggests increasing awareness of the importance of data diversity for robust model development, important gaps remain. In particular, demographic diversity is often insufficiently reported, limiting the assessment of model fairness. At the same time, dataset accessibility remains a critical bottleneck. Although more than half of the studies rely exclusively on public datasets, a significant proportion incorporate proprietary data or depend entirely on private collections. While private datasets may enhance scale and diversity, their use limits reproducibility and standardized benchmarking, underscoring the need for larger, standardized, and openly accessible radiology datasets \citep{kelly2019key,moor2023foundation}.

Regarding the second pillar, current radiology VFMs remain characterized by substantial diversity in architectural design, pretraining strategies, and computational scale, reflecting an evolving yet still fragmented methodological landscape.

Transformer-based architectures clearly predominated across studies, likely due to their ability to capture long-range spatial dependencies, integrate multi-scale contextual information, and scale effectively with increasing data availability \citep{azad2024advances}. However, the continued presence (>30\%) of purely convolutional and hybrid convolution–transformer architectures suggests that fully transformer-based solutions may not yet adequately capture the inductive biases required for radiological imaging data. In particular, convolutional components remain important for encoding fine-grained local structures, which are often critical for radiological interpretation \citep{kim2025systematic}. The coexistence of these paradigms indicates that the field has not yet converged on a canonical architectural design for VFMs in radiology. At the same time, the emergence of alternative approaches—including state-space models (e.g., Mamba \citep{mamba}), graph-based methods, and mixture-of-experts frameworks—highlights ongoing exploration beyond established transformer-based designs. Although these paradigms remain less prevalent, they point toward promising directions for addressing challenges such as spatio-temporal modeling, structured reasoning, modality-aware processing, and computational efficiency. 

Pretraining strategies similarly revealed a strong predominance of SSL, likely reflecting both the scarcity of large-scale annotated radiological datasets and the need to exploit large quantities of unlabeled medical imaging data for representation learning \citep{de2025foundation}. Within this context, reconstruction-based approaches, particularly masked image modeling, are the most commonly adopted SSL strategy, followed by contrastive learning and composite multi-stage pipelines, typically combining reconstruction and contrastive objectives \citep{wang2023review}. Emerging techniques such as self-distillation are also observed, likely reflecting their recent success in natural-image domains thanks to their capability to integrate both global- and pixel-level features \citep{dino}. Nevertheless, a non-negligible proportion of studies continues to rely on supervised pretraining, including multi-task and segmentation-oriented setups, particularly in settings where curated annotations were available. 

In parallel, the increasing use of initialization from large pretrained models (e.g., SAM-based approaches \citep{sam}) highlights a growing dependence on external sources of knowledge. While these strategies can improve efficiency and performance, they also raise questions regarding domain alignment and the extent to which general-domain representations adequately capture the specific characteristics of radiological data.

Model scale, computational requirements, and reproducibility practices also revealed valuable insights  in the current radiology VFMs landscape. Reported parameter counts span several orders of magnitude, ranging from relatively compact architectures to models approaching the billion-parameter scale, although the median size ($\sim$200M) remains substantially lower than that of FMs in natural vision or multimodal AI \citep{patil2024review}. This suggests that, despite adopting the terminology of FMs, current approaches in radiology are still comparatively limited in scale. This trend is further reflected in the reported computing infrastructure, as a limited number of studies reported access to large multi-GPU infrastructures, while a considerable proportion relied on single-GPU setups. These constraints likely stem from the restricted availability of large-scale training datasets, which in turn reduces the necessity of large-scale distributed training infrastructures. 

Reproducibility and openness practices presented a mixed picture. Although a majority of studies provided access to source code, substantially fewer released pretrained weights, thereby limiting reproducibility and downstream reuse. In addition, incomplete reporting of model configurations and computational resources in some works further hinders transparent comparison across methods. As models continue to grow in complexity, standardized reporting practices and broader adoption of open science principles will be essential to ensure reproducibility and facilitate progress in the field.

Regarding the third pillar, current radiology VFMs were primarily assessed in terms of the breadth of their downstream evaluation, the transferability of their learned representations, and their advantages over task-specific approaches.

Most studies (86.5\%) evaluated their models across multiple downstream tasks, reflecting an emphasis on assessing the transferability and general-purpose representation capabilities. However, the wide range in the number of tasks per study (1–146) and the relatively modest median (6) highlight a substantial heterogeneity in how scalability to multiple applications is demonstrated. Additionally, only a minority of studies assessed performance across more than ten distinct tasks, suggesting a relatively narrow experimental evaluation and limiting the strength of applicability conclusions.

The distribution of downstream tasks further reinforces this observation. Segmentation and classification remained the predominant evaluation settings, accounting for more than half of the reported downstream tasks. This comparatively limited exploration of more complex or longitudinal tasks—such as survival analysis, progression modeling, and image enhancement—suggests an incomplete assessment of VFMs’ capabilities to capture temporal evolution phenomena across diverse radiological scenarios. Broader real-world adoption will therefore require stronger evidence in these clinically meaningful settings.

Beyond task diversity, radiological data heterogeneity underscores the importance of explicitly evaluating representation transferability across distribution shifts. Generalization analysis was performed in nearly all studies, most commonly through evaluations under disease shift conditions (80.5\%), indicating a strong focus on robustness across varying disease distributions. In contrast, modality, and anatomical shifts were less frequently assessed, potentially reflecting the predominance of domain-specific FMs trained on restricted modalities and anatomical regions. Similarly, although multicenter datasets were commonly used during pretraining, explicit cross-center and cross-scanner evaluations remained comparatively limited, despite their recognized importance for assessing robustness under real-world clinical variability. Overall, VFMs demonstrate encouraging robustness within constrained settings, but their generalizability across heterogeneous clinical environments remains insufficiently characterized \citep{suleman2025assessing}.

From an adaptation perspective, full or partial fine-tuning remained the dominant strategy (67.1\%), often complemented by parameter-efficient fine-tuning (PEFT) approaches. This pattern indicates that most VFMs still rely heavily on task-specific parameter updates to achieve competitive performance. Linear probing, zero-shot, and few-shot evaluations were comparatively less explored, suggesting that current pretrained representations are not yet sufficiently generalizable to support robust out-of-the-box transfer across heterogeneous radiological settings, where labeled data may be scarce or unavailable. Even so, broader evaluation of these approaches may be valuable for reducing the dependence on annotated downstream datasets for fine-tuning, while also lowering adaptation costs, computational requirements, and time-to-deployment.

Performance comparisons generally suggest strong empirical results, with most studies reporting improvements over pretrained or competing FMs (93.8\%) and task-specific supervised models (84\%). However, these results should be interpreted cautiously. First, the absence of standardized benchmarking protocols and the heterogeneity of datasets, evaluation metrics, and experimental designs limit the comparability of results across studies. Second, potential biases related to dataset selection, overlap between pretraining and evaluation data, and insufficient external validation may contribute to optimistic estimates of performance.

The assessment of radiology VFMs using the FUTURE-AI principles \citep{futureaiguidelines} revealed that fewer than one-third of studies provided comprehensive coverage across all assessed dimensions. This pattern indicates that, despite increasing awareness of responsible AI frameworks, their systematic integration into model development and evaluation remains limited.

A clear trend emerges in the prioritization of technically driven principles. Robustness was the most consistently addressed dimension, largely reflecting the emphasis on generalization across datasets and downstream tasks. Similarly, the high prevalence of universality and usability aligns with the overarching goal of VFMs to function as general-purpose models adaptable to multiple applications. Traceability was also frequently reported, typically through documentation of model architectures, training procedures, and evaluation protocols. However, the depth and granularity of such reporting remain variable, and often emphasizing methodological transparency rather than full reproducibility through access to pretrained weights, datasets, or complete training pipelines. 

In contrast, explainability and fairness exhibit comparatively lower and more inconsistent coverage, revealing an important imbalance in the current research landscape. Explainability, when addressed, was often limited to post hoc visualization techniques (e.g., activation maps or saliency methods), which may provide constrained insight into the clinically meaningful factors underlying the model predictions. The reduced assessment of fairness is particularly notable, as only half of the studies explicitly considered this dimension. Fairness assessments were often restricted to basic subgroup analyses or remain implicitly addressed through dataset diversity, rather than through systematic bias quantification or mitigation strategies. 

Taken together, these findings suggest a certain misalignment between the current emphasis of VFM research on representation learning and performance optimization and the broader requirements for trustworthy clinical deployment. Addressing these gaps will require not only methodological advances but also standardized evaluation frameworks that integrate technical performance with ethical and clinical considerations. In this context, the FUTURE-AI framework provides a valuable reference for guiding the next phase of radiology VFM development.

\section{Limitations and further work}

This review has several limitations. First, our restriction to imaging-only models excluded multimodal architectures incorporating language encoders or clinical text. Although this choice preserved methodological comparability within a controlled scope, it limited the breadth of the synthesis and should be addressed in future work.

Second, although we employed a structured and intentionally broad search strategy, relevant studies may have been missed because VFM terminology remains recent and inconsistently used in radiology. Coverage may also have been limited by the exclusion of additional sources such as conference proceedings and preprint repositories, and periodic updates will be needed given the rapid methodological evolution of the field.

Third, the absence of universally accepted criteria for defining radiology VFMs introduced interpretative judgment into study selection. Although our conceptual framework was designed to establish structured inclusion boundaries, classification decisions inevitably involved some subjectivity, particularly in borderline cases.

Finally, as this work was conducted as a scoping review, we did not perform formal methodological quality assessment or quantitative synthesis of reported performance metrics. Future systematic reviews or meta-analyses may enable more rigorous evaluation of study quality, benchmarking practices, reproducibility standards, and comparative effectiveness across VFM approaches.
\section{Conclusion} 

This scoping review highlights the rapid emergence of VFMs in radiology while underscoring the early-stage and fragmented nature of the field. 
Despite encouraging progress, current radiological VFMs remain constrained by limited dataset scale, modality imbalance, and substantial methodological heterogeneity.

Across the three core pillars examined—data, methodology, and evaluation—a consistent theme is the absence of standardized practices. Pretraining datasets remain relatively small and insufficiently diverse, architectural and training strategies are highly variable, and downstream evaluation practices lack uniform benchmarks. 

Although studies report strong empirical performance, the field remains predominantly focused on technical optimization, with comparatively limited attention to trustworthiness dimensions such as fairness, explainability, and reproducibility. Strengthening alignment with responsible AI frameworks, including FUTURE-AI, will be critical for enabling trustworthy clinical translation.

Addressing these challenges will be essential to realize the full potential of VFMs as scalable, general-purpose tools for robust, equitable, and clinically meaningful medical image analysis.


\section*{Acknowledgments}
The authors declared no financial support was received for this work.

\section*{Author contributions}
A.V.-R. and X.R.-P. contributed equally to this work. A.V.-R., X.R.-P., and A.J.-P. conceived the study. A.V.-R., X.R.-P., and A.F.-M. designed the methodology. A.V.-R. and X.R.-P. developed and executed the search strategy, performed screening, data extraction, and data analysis, and drafted the original manuscript. A.F.-M., I.I.R., Á.A.-B., and A.J.-P. critically reviewed and edited the manuscript. A.J.-P. supervised the work. All authors read and approved the final manuscript.

\section*{Competing interests}
A.V.-R., X.R.-P., A.F.-M., I.I.R., and A.J.-P are employed by Quibim SL, a medical imaging technology company. Á.A.-B. is the CEO of Quibim SL and has stock ownership in the company.

\bibliographystyle{npj_nature_references}
\bibliography{bibliography}

\clearpage
\section{Supplementary Material}

\subsection{Search terms}
\label{sec:sm_1}
Search terms used to include and exclude studies. Inclusion terms are searched in title, abstract, and keywords, whereas only the title is used to exclude publications. Publication year, source type, and language were considered as well.

\begin{table}[h]
\centering
\caption{Literature search strategy and filtering criteria.}
\label{tab:search_strategy}
\setlength{\tabcolsep}{0pt}
\renewcommand{\arraystretch}{1.15}
\begin{tabular}{p{6.5cm} p{9.5cm}}
\toprule
\textbf{Category} & \textbf{Search terms / filters} \\
\midrule

\textbf{Inclusion (TITLE-ABS-KEY)} 
& \textbf{Foundation model} \\
& ``foundation* model*''; ``large vision model*''; ``large model*''; ``vision foundation model*''; ``VFM''; ``large pre?trained model*''; ``large ai''; ``universal model*''; ``general-purpose model*'' \\
& \textbf{AND} \\
& \textbf{Medical imaging} \\
& ``medical imag*''; ``radiolog*''; ``radiolog* imag*''; ``computed tomography''; ``CT''; ``magnetic resonance''; ``MRI''; ``x-ray''; ``nuclear medicine''; ``PET''; ``ultrasound''; ``clinical imag*''; ``oncolog*'' \\
\midrule

\textbf{Exclusion (TITLE)} 
& ``natural language''; ``large language model*''; ``fundus''; ``OCT''; ``pathology''; ``pathological''; ``electronic health''; ``EHR''; ``genomic''; ``text''; ``notes''; ``free-text''; ``language''; ``LLM''; ``gene''; ``genes''; ``genetic*''; ``genom*'' \\
\midrule

\textbf{Publication year} 
& $>$2016 and $<$ March 2026 \\
\midrule

\textbf{Source type} 
& Journals only (LIMIT-TO(SRCTYPE, ``j'')) \\
\midrule

\textbf{Language} 
& English (LIMIT-TO(LANGUAGE, ``English'')) \\
\bottomrule
\end{tabular}
\end{table}

\clearpage
\subsection{Search strategy}
\label{sec:sm_search_strategy}

Full electronic search strategy for Scopus is provided below.
\begin{Verbatim}[fontsize=\small]
(
  (
    TITLE-ABS-KEY("foundation* model*") OR
    TITLE-ABS-KEY("large vision model*") OR
    TITLE-ABS-KEY("large model*") OR
    TITLE-ABS-KEY("vision foundation model*") OR
    TITLE-ABS-KEY("VFM") OR
    TITLE-ABS-KEY("large pre?trained model*") OR
    TITLE-ABS-KEY("large ai") OR
    TITLE-ABS-KEY("universal model*") OR
    TITLE-ABS-KEY("general-purpose model*")
  )
  AND
  (
    TITLE-ABS-KEY("medical imag*") OR
    TITLE-ABS-KEY("radiolog*") OR
    TITLE-ABS-KEY("radiolog* imag*") OR
    TITLE-ABS-KEY("computed tomography") OR
    TITLE-ABS-KEY("CT") OR
    TITLE-ABS-KEY("magnetic resonance") OR
    TITLE-ABS-KEY("MRI") OR
    TITLE-ABS-KEY("x-ray") OR
    TITLE-ABS-KEY("nuclear medicine") OR
    TITLE-ABS-KEY("PET") OR
    TITLE-ABS-KEY("ultrasound") OR
    TITLE-ABS-KEY("clinical imag*") OR
    TITLE-ABS-KEY("oncolog*")
  )
)
AND NOT
(
  TITLE("natural language") OR
  TITLE("large language model*") OR
  TITLE("fundus") OR
  TITLE("OCT") OR
  TITLE("pathology") OR
  TITLE("pathological") OR
  TITLE("electronic health") OR
  TITLE("EHR") OR
  TITLE("genomic") OR
  TITLE("text") OR
  TITLE("notes") OR
  TITLE("free-text") OR
  TITLE("language") OR
  TITLE("LLM") OR
  TITLE("gene") OR
  TITLE("genes") OR
  TITLE("genetic*") OR
  TITLE("genom*")
)
AND PUBYEAR > 2016 AND PUBYEAR < Msarch 2026
AND (LIMIT-TO(SRCTYPE, "j"))
AND (LIMIT-TO(LANGUAGE, "English"))
\end{Verbatim}

\clearpage
\subsection{FUTURE-AI Principles Reporting}
\label{sec:sm:futureai}
\begin{landscape}
\scriptsize
\setlength{\tabcolsep}{3pt}
\renewcommand{\arraystretch}{1.1}

\newcolumntype{P}[1]{>{\raggedright\arraybackslash\setlength{\baselineskip}{0.75\baselineskip}}p{#1}}

\rowcolors{2}{gray!5}{gray!15}

\begin{longtable}{P{1cm} P{3.5cm} P{3.5cm} P{3.5cm} P{3.5cm} P{3.5cm} P{3.5cm}}
\caption{FUTURE-AI principles reporting.}
\label{sm:tab:futureai}\\

\toprule
\rowcolor{white}
\textbf{Study} & \textbf{Fairness} & \textbf{Universality} & \textbf{Traceability} & \textbf{Usability} & \textbf{Robustness} & \textbf{Explainability}\\
\midrule
\endfirsthead

\hiderowcolors
\toprule
\rowcolor{white}
\textbf{Study} & \textbf{Fairness} & \textbf{Universality} & \textbf{Traceability} & \textbf{Usability} & \textbf{Robustness} & \textbf{Explainability}\\
\midrule
\endhead

\hiderowcolors
\midrule
\multicolumn{7}{r}{\emph{Continues in next page}}\\
\bottomrule
\endfoot

\hiderowcolors
\bottomrule
\endlastfoot

\showrowcolors

\cite{yu2024} & Not reported & Not reported & Code and model weights publicly available & Not reported & Evaluated on two independent lung CT datasets (COPDGene, MosMed) & 3D emphysema mask representations; metric evolution during fine-tuning\\
\cite{tang2025} & Multi-dataset, multi-organ training; bias acknowledged but not analyzed & Transfer across CT and MRI segmentation tasks & Public data sources and code available & Not reported & Consistent gains across multiple downstream datasets & Not reported\\
\cite{bhatt2023} & Not reported & Not reported & Not reported & Not reported & Not reported & Not reported\\
\cite{pai2024} & Not reported & Validated across multiple datasets and institutions & Open data, code, and pretrained weights & Containerized application provided & High test--retest ICC; stability under perturbations & Gradient-based saliency maps; gene-expression analyses\\
\cite{wang2025} & Dataset bias acknowledged across hospitals and demographics & Evaluated on English and Chinese report generation tasks & Public code available & Not reported & Multi-dataset evaluation with ablation studies & Activation response maps highlighting thoracic regions\\
\cite{yang2024} & Multi-center, multi-disease cohort with site-aware splits & Generalization across internal/external datasets; zero/few-shot & Detailed data sources, preprocessing, and training documentation & Pretrained weights and code released & External validation; sensitivity analyses & Attention maps and multivariate disease-region analyses\\
\cite{han2025} & Multiple datasets used & Not reported & Not reported & Not reported & Not reported & Visualization of key brain regions and relationships\\
\cite{he2024} & Large public datasets with diverse subjects & Zero-shot evaluation & Open-source code available & Not reported & Zero-shot evaluation on two datasets & Phenotype Active Maps linking regions to phenotypes\\
\cite{schafer2024} & Not reported & Multi-modality and multi-label training & Not reported & Frozen encoder enables efficient downstream training & External multi-center, multi-scanner validation & Not reported\\
\cite{alhammuri2025} & Human-in-the-loop annotation quality control & Not reported & Not reported & Semi-supervised workflow with human correction & Handles noisy data and missing inputs & Interpretable embeddings; QC metrics\\
\cite{chu2025} & Multi-source public and in-house datasets; external test sets & Applicable across nine modalities (2D/3D) & Ablation studies of architectural components & Lightweight decoder; efficient integration & Robust to perturbations; unseen modality generalization & Reconstruction and frequency-domain visualizations\\
\cite{silva2025} & Not reported & Cross-dataset transfer; adaptation to novel classes & Open data, code, and pretrained weights & Few-shot and parameter-efficient adaptation & Domain-shift and low-shot evaluation & Not reported\\
\cite{ding2025} & Broad multi-dataset, multi-disease training & Multicentric external validation & Architectural ablation studies & Pretrained encoder reuse discussed & External validation on held-out dataset & GradCAMs; expert-frequency analyses\\
\cite{gao2025} & Multiple imaging manufacturers included & Not reported & Not reported & Not reported & Not reported & Occlusion sensitivity mapping\\
\cite{wood2024} & Clinically representative, demographically diverse data & Transfer across MRI sequences and orientations & Detailed methods and open scripts & Pretrained models and fine-tuning scripts & Improved out-of-sample performance; variance reduction & Not reported\\
\cite{cui2023} & Large healthy and clinical datasets; imbalance acknowledged & Adaptability across ADHD, ASD, and PTE tasks & Ablation studies & Zero-shot inference via linear probing & Validated across multiple datasets & Feature-importance maps from classifier coefficients\\
\cite{suo2025} & Single-center, all-Asian cohort; scanner diversity described & Routine clinical MRI protocols for real-world applicability & Ethics approvals; code and pretrained models released & Minimal preprocessing; whole-brain inputs & Multi-scanner evaluation; independent test set & Occlusion sensitivity maps; voxel-wise statistics\\
\cite{shen2025} & Not reported & CT/MRI evaluation across organs and datasets & Not reported & Interactive and auto-prompting modes & Consistent performance across modalities; zero-shot classes & Prototype-based interpretable mask embeddings\\
\cite{gong2025} & Three-center evaluation & Multi-center protocols with heterogeneous image quality & Ethics approval; no shared code or weights & Integrated into clinical AI system & Ensemble modeling; external validation & GradCAMs; center-wise performance analysis\\
\cite{chen2024masam} & Not reported & Cross-modality evaluation (CT, MRI, surgical video) & Not reported & Promptable segmentation for semi-automatic workflows & External generalization; component ablations & Not reported\\
\cite{lin2025samct} & Not reported & Generalization across 30 datasets and 118 objects & Code, data, and checkpoints released & Automated task-indicator prompts & Strong performance on unseen datasets & Not reported\\
\cite{lin2025perceptguide} & Not reported & External validation across organs and datasets & Public code and pretrained models & Fully automated prompts & Balanced sampling; stable performance across datasets & Not reported\\
\cite{zhang2025} & Not reported & Task- and modality-agnostic design; missing-modality handling & Code repository available & Single universal model simplifies deployment & Generalization to unseen datasets; stability ablations & Expert probability maps; latent space visualizations\\
\cite{cox2024} & Large multi-cohort population with standardized preprocessing & Adaptable across tasks and MRI modalities & Clear dual-phase pretraining pipeline & Few-shot and modality-restricted training support & Robust under few-shot and restricted-modality settings & Ablations on modality restriction and few-shot learning\\
\cite{yang2025} & No subgroup or demographic bias analysis & Multi-modal and non-image extension capability & Public code (no pretrained weights) & Code reusable for downstream tasks & External validation on OASIS-3 & Not reported\\
\cite{zhou2025} & Not reported & Limited to four datasets; no cross-modality evaluation & No model weights released & Requires bounding-box input & Performance degrades on noisy/low-quality data & Not reported\\
\cite{chen2025mofo} & Multi-center, multi-vendor data with leakage-reducing splits & Multi-organ OOD evaluation & Open-source code and model & Fully automatic segmentation & Improved performance on challenging OOD datasets & Inter-organ relationship and latent-space analyses\\
\cite{sun2025} & Not reported & Tested on 19 datasets across lifespan and MRI contrasts & No code or weights shared & Not reported & Robustness quantified across simulated artefacts & Not reported\\
\cite{gu2025} & Organ distribution and demographic analysis reported & External validation across locations and sequences & Code and weights available & Interactive prompt-based refinement & Generalization to unseen locations and artefacts & Not reported\\
\cite{chen2025} & Multi-center, multi-vendor, multi-pathology evaluation & Generalization to unseen vendors and centers & Public code; IRB approvals & Text/box prompts; near-real-time inference & Strong external performance with stratified analyses & Not reported\\
\cite{gu2024} & Broad dataset; fairness not explicitly analyzed & Extended external evaluation & Not open-source & Interactive segmentation & External validation on 12 multi-modal datasets & Not reported\\
\cite{lei2025} & Diverse public CT datasets; no demographic analysis & General-purpose 3D localization and segmentation & Open datasets, code, and models & Minimal manual input via automated prompts & Strong cross-dataset generalization & Interpretable 3D bounding boxes and spatial maps\\
\cite{ma2025} & Sex-bias tolerance evaluated across splits & Zero-shot transfer across tasks and datasets & Data, code, and models public & Fully open fine-tuning and adaptation & Robust under long-tailed and domain-shift conditions & Not reported\\
\cite{machado2025} & Not reported & Multi-organ training; multicentric external validation & Not open-source & Radiologist-in-the-loop interactive prompts & External real-world validation; editing mitigates bias & Visual prompts and interactive editing\\
\cite{jiao2024} & Organ-balanced sampling strategy analyzed & Multi-organ, multi-center, multi-device evaluation & Open-source code and weights & Plug-and-play backbone integration & Robust feature learning under low-quality US conditions & Not reported\\
\cite{sun2024} & Not reported & Cross-domain generalization across multiple modalities & Code and models not public & Automated prompt-free segmentation & Robust across modality variation & Not reported\\
\cite{kang2024} & Single-center, single-organ cohort & Transfer across vendors and tasks & Code and weights available & Open-source model & Improved robustness to image blur; cross-device testing & Not reported\\
\cite{luo2025} & Subgroup analyses by age, BPE, BI-RADS, field strength & Multi-center, multi-protocol external validation & IRB approvals; code released & Clinical decision-curve analyses & External validation; missing-modality robustness & Integrated gradients; Shapley values\\
\cite{chen2024brain} & Large multi-scanner dataset; external TCGA validation & Multi-scanner, multi-task evaluation & Not reported & Minimal preprocessing; whole-brain input & Robust to adversarial and noise perturbations & Occlusion-based saliency maps\\
\cite{ma2024} & Large multi-modality public dataset; no subgroup analysis & Single model across organs, modalities, and diseases & Open data, code, and weights & Prompt-based segmentation reduces annotation time & Extensive internal and external validation & Mask confidence scores provided\\

\end{longtable}
\end{landscape}

\end{document}